\documentclass[preprint,10pt,3p]{elsarticle}
\usepackage{geometry}
\usepackage{booktabs}
\usepackage{tabularx}
\usepackage{graphicx}
\usepackage{multirow}
\usepackage{caption}
\usepackage{amssymb}
\usepackage{amsmath}
\usepackage{algorithm}
\usepackage{algpseudocode}
\usepackage{amsfonts}
\usepackage{subfigure}
\usepackage{times}
\usepackage{xcolor}
\usepackage[colorlinks=false, linkcolor=black, citecolor=black, urlcolor=black]{hyperref}

\begin{document}

\begin{frontmatter}

\title{Traffic expertise meets residual RL: Knowledge-informed model-based \\ residual reinforcement learning for CAV trajectory control}

\author{Zihao Sheng}
\author{Zilin Huang}
\author{Sikai Chen\corref{cor}}
\ead{sikai.chen@wisc.edu}
\cortext[cor]{Corresponding author: Sikai Chen}

\address{Department of Civil and Environmental Engineering, University of Wisconsin-Madison, Madison, WI, 53706, USA}

\begin{abstract}
Model-based reinforcement learning (RL) is anticipated to exhibit higher sample efficiency compared to model-free RL by utilizing a virtual environment model. However, it is challenging to obtain sufficiently accurate representations of the environmental dynamics due to uncertainties in complex systems and environments. An inaccurate environment model may degrade the sample efficiency and performance of model-based RL. Furthermore, while model-based RL can improve sample efficiency, it often still requires substantial training time to learn from scratch, potentially limiting its advantages over model-free approaches. To address these challenges, this paper introduces a knowledge-informed model-based residual reinforcement learning framework aimed at enhancing learning efficiency by infusing established expert knowledge into the learning process and avoiding the issue of beginning from zero. Our approach integrates traffic expert knowledge into a virtual environment model, employing the Intelligent Driver Model (IDM) for basic dynamics and neural networks for residual dynamics, thus ensuring adaptability to complex scenarios. We propose a novel strategy that combines traditional control methods with residual RL, facilitating efficient learning and policy optimization without the need to learn from scratch. The proposed approach is applied to CAV trajectory control tasks for the dissipation of stop-and-go waves in mixed traffic flow. Experimental results demonstrate that our proposed approach enables the CAV agent to achieve superior performance in trajectory control compared to the baseline agents in terms of sample efficiency, traffic flow smoothness and traffic mobility. 
The source code and supplementary materials are available at: \href{https://zihaosheng.github.io/traffic-expertise-RL/}{\textcolor{magenta}{https://zihaosheng.github.io/traffic-expertise-RL/}}. 
\end{abstract}


\end{frontmatter}

\section{Introduction}

Connected automated vehicles (CAVs) represent a pivotal innovation in the evolution of future transportation networks \citep{Liu2023Can,Qu2023Envisioning,Liao2024GPT,Sheng2024Kinematics}. By harnessing advanced technologies like sensors, artificial intelligence, and communication systems, CAVs hold the promise of significantly enhancing road safety, travel efficiency, and environmental sustainability \citep{Olovsson2022Future,Dong2022Development,Dong2023Why,Chen2023Taxonomy,Huang2024Toward}. This potential has fueled an increasing focus on the development of safe and intelligent control strategies for CAVs. Such strategies are designed to enable these vehicles to function as dynamic actuators within mixed traffic, thereby optimizing traffic flow and reducing congestion \citep{Stern2018Dissipation,Cui2017Stabilizing,Andreotti2023Potential}. Classical optimization-based approaches have achieved considerable successes in this context, but these approaches often encounter limitations due to their dependence on the careful selection and tuning of parameters \citep{Shi2023deep,Garriga2010Model}. This process can be both labor-intensive and specific to each application, thereby struggling to maintain performance in the face of unexpected or unusual conditions.

Learning-based approaches have shown a remarkable ability to adapt and thus overcome the limitations inherent in the above classical strategies \citep{Ding2022enhanced,Kiran2022Deep,Sheng2024Ego}. Within the spectrum of learning-based approaches, reinforcement learning (RL) has emerged as a promising solution for handling complex traffic control tasks \citep{Li2023survey,Han2023Leveraging,Du2023Dynamic}. RL aims to learn an optimal control strategy through interactions between an agent and the environment. Existing RL techniques are primarily model-free, which means they do not utilize prior knowledge about environmental dynamics but instead learn optimal actions through direct interaction with the actual environment. Model-free RL has proven effective in tackling a wide range of tasks, such as robot control and autonomous driving, demonstrating exceptional capabilities \citep{Haarnoja2018Soft}. However, model-free RL is often criticized for its low sample efficiency and slow convergence speed. This becomes particularly problematic for complex tasks, where training an RL agent from scratch requires vast amounts of interaction with the actual environment and can be time consuming. Additionally, poor initialization can occasionally lead to premature convergence at local minima. These inherent limitations prevent the practical implementation of RL in traffic control tasks, because real-world interactions within a traffic system are usually highly expensive and dangerous.

Model-based RL has emerged as an innovative solution to overcome the limitations of model-free RL by incorporating a virtual environment model to augment the data obtained from actual-environment interactions \citep{Moerland2023Model}. This strategy enables model-based RL to utilize virtual models for predicting state transitions, thus simulating agent interactions in a controlled setting. As a result, model-based RL not only enhances sample efficiency and accelerates the learning process but also reduces the risks of encountering hazardous situations during training. Nonetheless, the construction of an accurate virtual environment model presents a significant challenge within model-based RL. Specifically, constructing models that accurately reflect the complexity of real-world dynamics often requires extensive domain-specific expertise and a deep understanding of the underlying system dynamics, which is not always readily available \citep{Wu2023Uncertainty,Chee2022KNODE}. Neural networks (NNs) have shown promise in learning dynamic models from data, but they often suffer from performance degradation in closed-loop RL settings. Moreover, despite the advancements in model-based RL, there is still a challenge that remains unaddressed: the necessity for the agent to initiate learning from scratch. This starting point means that despite utilizing virtual models for enhanced learning efficiency, model-based RL agents must initially navigate through a learning curve similar to their model-free RL counterparts.

To address the above limitations, this paper proposes a knowledge-informed model-based residual reinforcement learning framework that enhances learning efficiency by leveraging established knowledge. Instead of starting from scratch, it is more efficient to learn residuals based on guidance derived from existing well-established knowledge, such as physical principles and optimization algorithms, for both the virtual environment model and the RL agent. Firstly, we propose a virtual environment model informed by expert knowledge, which leverages the Intelligent Driver Model (IDM) \citep{Treiber2000Congested} to provide insights into the average dynamics of the traffic system. To complement this, NNs are incorporated to capture and adjust for the residual and uncertain dynamics within the system, aiming to enhance the adaptability to complex real-world scenarios. Secondly, in addressing the complex tasks of traffic flow control, our framework adopts a novel approach by combining a conventional suboptimal but stable controller with an overlay of a residual RL agent. The conventional controller is regarded as an initial policy, and the residual RL agent acts as a corrective term to fine-tune the policy towards perfection. Based on that, we develop a knowledge-informed model-based residual reinforcement learning framework, aiming to enhance the capabilities of imperfect controllers and to enhance the data efficiency of RL agents. The proposed framework is applied to CAV control for smoothing stop-and-go traffic flow, demonstrating its capability to effectively manage and mitigate traffic disturbances. The main contributions of this work are as follows:

1. We propose a novel framework, named knowledge-informed model-based residual reinforcement learning, which integrates well-established domain knowledge into the model-based RL learning process, encompassing both the virtual environment model and the RL agent. By leveraging domain expertise in transportation science and combining it with the adaptability of neural networks and the efficiency of residual learning, our framework substantially improves upon existing approaches in addressing the challenges of sample inefficiency, the need to learn from scratch, and the difficulty in accurately modeling complex environment dynamics. We provide a comprehensive theoretical analysis, including proofs of convergence and performance bounds, to rigorously demonstrate the advantages of our approach over existing methods.

2. The framework features a unique virtual environment model informed by expert knowledge, specifically leveraging the IDM to describe basic dynamics and neural networks to capture residual dynamics. This hybrid approach allows the model to maintain a foundational understanding of traffic flow theory while adapting to complex scenarios that may deviate from idealized conditions. The IDM provides a reliable baseline for vehicle behavior, while the neural network component learns to adjust for real-world complexities and individual driving variations not captured by the IDM. This design overcomes the low sample efficiency issues associated with model-free RL by simulating real-world interactions within a controlled virtual setting, while also providing more accurate and adaptable simulations compared to purely model-based approaches.

3. Our approach advances a residual reinforcement learning strategy that harmonizes the reliability of traditional control laws with the dynamic adaptability of a residual RL agent. Instead of learning from zero, this innovative method optimizes a residual policy using RL algorithms, based on an initial expert policy derived from existing controllers. The expert controller establishes a foundation of basic performance and reliability, while the residual RL agent unlocks the potential for continual learning and improvement. The residual RL agent learns to refine and improve upon this initial policy, focusing on aspects of the control problem that are not adequately addressed by the traditional controller. By decomposing the learning problem into a known component (provided by expert knowledge) and an unknown residual, we simplify the learning process and accelerate convergence to optimal policies. 

4. We provide a comprehensive theoretical analysis that rigorously demonstrates the advantages of our approach. This analysis includes proof of convergence, bounds on performance improvements, and explanations of why our integrated approach leads to more efficient and effective learning in mixed traffic environments. Specifically, we present three key theorems: (1) demonstrating that our physics-based initial policy provides a better starting point compared to random initialization, (2) proving the convergence properties of our proposed policy to an $\epsilon$-optimal policy, and (3) establishing the effectiveness of our virtual environment model in policy optimization. This theoretical foundation not only supports our claims of novelty but also provides valuable insights into the fundamental principles underlying our method’s effectiveness.

The remainder of this paper is organized as follows. Section \ref{sec2} reviews related works. Then, the proposed approach is illustrated in Section \ref{sec3}. Section \ref{sec4} introduces the implementation detail and experiment results. Finally, conclusions are drawn in Section \ref{sec5}. 

\section{Related works}\label{sec2}
\subsection{Classical analytical controller}
Enhancing driving behavior and performance through automated driving techniques unlocks promising avenues for achieving smoother traffic flow and heightened mobility efficiency \citep{Cui2017Stabilizing,Zheng2020Smoothing,Yue2022Effects}. Notably, real-world experiments by \citet{Stern2018Dissipation} have shown promising potential of such technologies in mitigating stop-and-go traffic patterns. Among these technologies, Adaptive Cruise Control (ACC) is a typical and widely used system known for its ability to autonomously adjust vehicle speeds by detecting the relative distance and speed to the leading vehicle \citep{Vahidi2003Research}. 
The introduction of vehicle-to-vehicle or vehicle-to-infrastructure wireless communication in Cooperative ACC (CACC) further augments this capability by leveraging additional information from multiple nearby vehicles, and significantly improves string stability and mobility \citep{Milanes2014Cooperative}. From a control perspective, the implementation of these technologies has largely relied on feedback and feedforward controllers, including proportional-integral-derivative (PID) control \citep{Lidstrom2012Modular}, and model-based control strategies \citep{Yu2022Eco}. These model-based control usually utilizes car-following models, such as Newell model \citep{Newell2002simplified}, and the Intelligent Driver Model \citep{Treiber2000Congested}. Leveraging these foundational models, a variety of model-based control strategies have been developed, ranging from optimal control and H-infinity control \citep{Zheng2020Smoothing,Zhou2020Stabilizing} to model predictive control (MPC) \citep{Feng2021Robust}. Additionally, the seminal works of \citet{Zheng2020Smoothing} and \citet{Cui2017Stabilizing} provide theoretical proofs and experimental evidence demonstrating that a strategically controlled autonomous vehicle can act as a mobile actuator in traffic flow, effectively damping stop-and-go waves and improving overall traffic stability.

MPC stands out as a predominant tool in tackling traffic control tasks with constraints, with a focus on the optimization of multiple objectives \citep{Feng2021Robust,Sheng2023Cooperation}. The essence of MPC lies in its ability to predict and optimize the future behavior of a system based on a system dynamics model in a rolling/receding horizon manner. 
Typically, MPC frameworks are designed with vehicle velocities and positions as the state variables, employing acceleration as the control variables to achieve these objectives \citep{Li2023survey}. For instance, \citet{Yang2023eco} adopted MPC to regulate the acceleration and velocities of CAVs for minimized fuel consumption, taking into account the dynamics of surrounding vehicles. Beyond minimizing fuel consumption, the objective function of MPC controllers can encompass a broader spectrum of considerations, including reducing travelling delay \citep{Gong2018Cooperative}, and improving driving safety \citep{Wang2024Collision}. 
Nevertheless, one of the primary challenges with MPC is its intensive computational requirements. The need to continuously predict future states and solve optimization problems over a receding horizon can be computationally burdensome \citep{Shi2023deep}. Moreover, the effectiveness of MPC heavily relies on the predetermined parameters within its predictive model and objective function, including weights for different objectives (e.g., safety, efficiency, comfort) and model coefficients that describe vehicle dynamics. The choice of these parameters can significantly affect the performance of the control system, requiring careful tuning and calibration \citep{Garriga2010Model}.

\subsection{Reinforcement learning}
RL-based approaches have recently emerged as powerful tools in addressing complex driving tasks \citep{Wu2022Intersection,Chen2021Graph,Wang2023GOPS}. This methodology involves an agent that learns optimal decision-making through interactions with its environment, aimed at maximizing cumulative rewards. The strengths of RL-based approaches are primarily manifested across three dimensions. 
First, this learning paradigm offers a significant advantage over classical analytical controllers by enabling agents to learn behaviors via trial and error, thus equipping them to adapt to new and unforeseen scenarios \citep{Huang2024Human}. 
Second, its computational intensity is predominantly confined to the offline training phase. This allows the trained driving policies to be executed swiftly in real-time applications, addressing the limitation of extensive computation times often associated with MPC \citep{Shi2023deep,Liu2024distributed}. 
Third, unlike conventional control methods that require to define objective functions explicitly through control actions, RL can implicitly encode them within the reward function. This indirect approach to specifying control goals enables RL to adeptly handle tasks that are challenging for conventional analytical controllers \citep{Staessens2022Adaptive}. 
Building on these advantages, RL-based approaches have shown remarkable success across a broad spectrum of challenging driving scenarios, including ramp merging \citep{Kiran2022Deep,Zhu2022Merging}, bottleneck management \citep{Ha2023Leveraging}, and intersection navigation \citep{Peng2021Connected,Liu2023Longitudinal}. For instance, \citet{Guo2023CoTV} demonstrated the superiority of PPO-based approaches in their cooperative traffic light and vehicle control system. Their study showed that PPO-based controllers achieved up to 30\% reduction in both travel time and fuel consumption and emissions under varying CAV penetration rates, outperforming traditional control methods. Similarly, \citet{Yang2024Eco} showcased the effectiveness of PPO and SAC in optimizing eco-driving strategies for mixed traffic near signalized intersections. Their results indicated that RL-based methods, particularly SAC, outperformed human drivers in all aspects and showed better energy efficiency compared to traditional models like IDM. However, the majority of these methods are model-free RL and often suffer from low sample efficiency due to the necessity of training the RL agent from scratch, resulting in substantial computational resource usage and prolonged training durations.

Model-based RL has gained significant attention in recent years as a powerful paradigm to address the sample inefficiency and exploration challenges in model-free RL \citep{Moerland2023Model}. Within the realm of traffic management, model-based RL introduces innovative strategies for addressing the complex transportation challenges. One of the pioneering works in model-based RL is the Dyna-style algorithm \citep{Sutton1990Integrated}, which utilizes collected interaction samples to learn a virtual model of the environment. This model allows the agent to generate simulated experiences, thereby facilitating the update of its policy based on these synthetic interactions. For instance, \citet{Yavas2022Model} have leveraged a Dyna-style model-based meta-policy optimization approach to ensure safer and more comfortable car-following behaviors. Similarly, \citet{Lee2020Model} proposed a Dyna-based model-based RL to optimize the control of electric vehicles, aiming for energy-efficient eco-driving. Furthermore, \citet{Huang2021ModelLight} introduced meta learning techniques into a model-based RL framework tailored for traffic signal control, with a focus on enhancing travel efficiency. Nevertheless, the RL agent within Dyna-style model-based RL still requires learning from scratch, necessitating extensive training periods to achieve optimal performance.

Another branch of model-based RL integrates the principles of MPC. This approach often employs a learned model of the environment to predict future states and rewards. It then solves the optimal sequence of actions over a defined time horizon using these predictions in an MPC manner \citep{Chua2018Deep}. For instance, \citet{Pan2021Integrated} introduced an MPC-based RL algorithm tailored for the control of mixed freeway traffic, utilizing a gradient-free cross-entropy technique to solve the optimal action sequence. 
In a similar vein, \citet{Dong2021Addressing} developed an MPC-based RL strategy for CAVs, specifically designed to prevent collisions in imminent crash scenarios by applying a random shooting method to calculate the optimal control actions. 
However, similar to MPC, MPC-based model-based RL also involves solving optimization problems at each time step to find the optimal action sequence, which can be computationally expensive. 
In the context of real-time CAV operation tasks, such computational intensity may hinder its deployment. Furthermore, it is worth noting that the overall performance of model-based RL methods is significantly influenced by the precision of the learned environment model.

\subsection{Residual RL}
The concept of learning residual models and policies falls into a broader framework of physics-informed machine learning \citep{Karniadakis2021Physics}, which synergizes the deductive and interpretable power of analytical physics-based models with the generalization and learning capabilities of data-driven models. 
Within the established frameworks, the integration of physics knowledge with NN models is accomplished in three distinct manners: generating supplementary data from physics models to enrich training datasets \citep{Han2022physics}, embedding constraints derived from physics into the loss functions of NN models \citep{Shi2022Physics}, and crafting specialized NN architectures that inherently incorporate prior knowledge \citep{Long2024Physics}. 
Residual model and policy learning, fitting into this third category, leverages the foundational performance offered by analytical physics-based models to account for basic and known dynamics, with data-driven models layered on top to address the residuals of unknown dynamics. 
The focus of residual model learning has predominantly been on predicting residuals to improve the accuracy of physics models, which has shown considerable promise in control tasks, ranging from autonomous racing \citep{Kabzan2019Learning} to robot control \citep{OConnell2022Neural}. 
Notably, in the field of transportation, the pioneering work of \citet{Long2024Physics} introduces the concept of physics-enhanced residual learning (PERL) for car-following trajectory prediction, marking a significant advancement in the application of residual model learning. 
Recent studies have further explored the integration of physics-based models with machine learning techniques in transportation scenarios. For instance, \citet{Chen2023Deep} employed IDM to predict HDV trajectories for collision risk assessment in multi-agent RL highway on-ramp merging, while \citet{Sun2023Cooperative} used IDM to generate expert demonstrations representing ideal car-following behavior for initializing their RL model through imitation learning. Additionally, \citet{Zhou2023Improving} proposed incorporating domain control knowledge into deep reinforcement learning agents for perimeter metering control, demonstrating improved learning efficiency, control performance, and scalability in urban network traffic management. 
However, these approaches rely heavily on physics models which may not fully capture the complexities of real-world traffic dynamics. As a result, the RL agents trained using these methods may inherit the inherent simplifications and biases of these physics models, potentially limiting their ability to adapt to and effectively handle more complex traffic scenarios encountered in real-world environments.

Distinct from residual models which primarily enhance prediction accuracy, residual policy learning decomposes a complex control task into two parts: one part that can be partially handled by conventional controllers for a near-optimal solution, and a residual part where RL techniques come into play \citep{Johannink2019Residual}. 
This division not only streamlines the learning process by leveraging existing controllers to bypass the initial learning from scratch but also enables continuous enhancement through RL-driven NNs. Recently, residual policy learning has captured attention across various domains. 
For instance, \citet{Hou2023Vehicle} leveraged residual policy learning to improve the braking controller for better passenger comfort during the post-braking process. 
\citet{Zhang2022Residual} designed a controller that integrates a modified artificial potential field with residual policy learning for autonomous racing, attaining performance on par with professional human racers across F1Tenth tracks. Furthermore, its applicability has extended to robot arm control \citep{Johannink2019Residual}, and parking scheduling \citep{Hou2024Hybrid}. 
Inspired by the efficiency and adaptability of residual learning, we aim to combine the advantages of both residual model learning and residual policy learning to effectively tackle the existing issues of low sample efficiency in model-free RL and the limitation of needing to learn from scratch in model-based RL. By leveraging the foundational insights provided by physics-based models and the dynamic adaptation facilitated by NNs within the RL framework, our approach promises to enhance sample efficacy, reduce computational demands, and expedite the training process.

\section{Methodology}\label{sec3}
\subsection{Overall framework}
\begin{figure*}
\centering
  \includegraphics[width=0.9\textwidth]{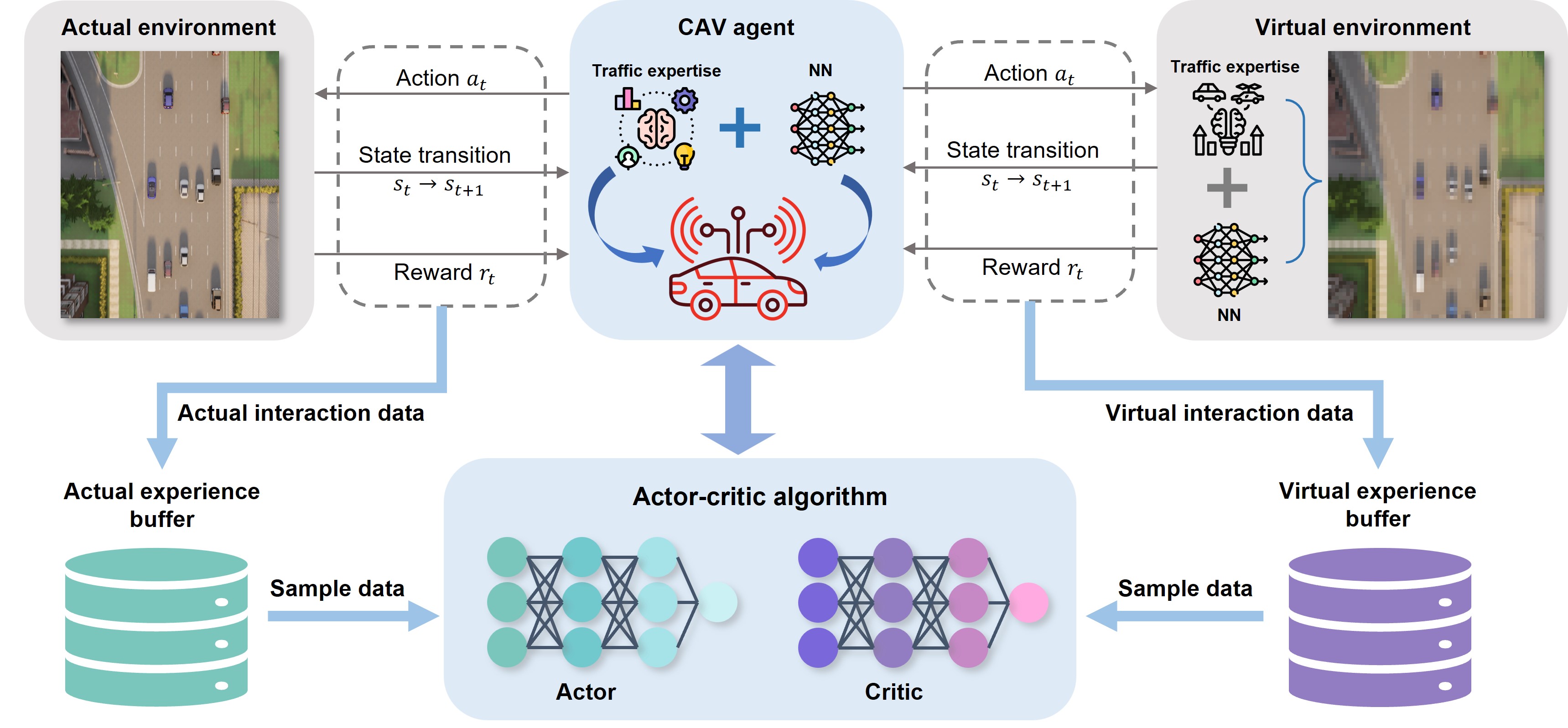}
  \caption{The overall framework of knowledge-informed model-based residual reinforcement learning.}
  \label{fig1}
\end{figure*}

Figure \ref{fig1} illustrates the overall framework of our proposed knowledge-informed model-based residual reinforcement learning. The CAV agent interacts with the actual traffic environment, and the interaction data are stored in the actual experience buffer. The CAV agent is uniquely designed to integrate existing traffic domain knowledge and neural networks (NNs). The traffic expertise serves as an initial policy to provide stable but rough actions, while the NNs continuously learn a residual policy to refine and optimize the actions provided by the traffic expertise. By incorporating prior knowledge, the CAV agent acquires a foundational understanding of the environment at the beginning of the training process, rather than learning from scratch with only a randomly initialized NN.

To further enhance the learning efficiency, we design a knowledge-informed virtual environment model to augment the interaction data for the update of the CAV agent. This virtual environment model leverages traffic expert knowledge to describe the basic and average traffic dynamics, and employs NNs to model and adapt to the residual and uncertain dynamics. The virtual environment model initiates its prediction from a state sampled from the actual experience buffer and then interacts with the CAV agent iteratively to generate virtual interaction data, which is stored in the virtual experience buffer. Both replay buffers are updated at every time step. During the training process, data from both actual and virtual experience buffers are sampled to update the CAV agent following an actor-critic algorithm. By exploiting the virtual environment model, the learning process of the RL agent can be significantly accelerated, requiring fewer interactions with the actual environment. It is important to note that our framework is not tied to any specific RL algorithm, physical model, neural network architecture, or simulators. Rather, it provides a flexible framework that can be adapted to various base models, learning algorithms, and simulator environments. The theoretical foundations of our method are detailed in \ref{appda}. These analyses provide proofs of convergence, performance advantages, and effectiveness of our approach.

\subsection{Virtual environment modeling with residual model}
In RL, the environment represents the actual system with which the RL agent interacts, defined by its states, actions, and rewards. At each time step $t$, the agent observes the current state $s_t$, takes an action $a_t$ based on its learned policy, and receives a reward $r_t$ as feedback from the environment. The environment then transitions to a new state $s_{t+1}$. The transition dynamics of the environment can be described by the following equation:
\begin{equation}
s_{t+1},r_t=T(s_t,a_t )
\end{equation}
To facilitate efficient exploration and learning for the RL agent while minimizing direct impact on the actual environment, we introduce a virtual environment. In this virtual environment, a model is used to approximate the transition function. The transition equation for the virtual environment is given by:
\begin{equation}
\hat{s}_{t+1}, \hat{r}_t = \hat{T}_{\psi}(s_t,a_t )
\end{equation}
where $\hat{s}_{t+1}$ and $\hat{r}_{t}$ denote the predicted state and reward in the virtual environment, respectively, and $\hat{T}_{\psi}(\cdot)$ represents the virtual transition function with parameter $\psi$, which can be learned from data or pre-defined based on domain knowledge. However, the virtual transition function often exhibits discrepancies between its predictions and the actual dynamics, which can be captured by a residual term:
\begin{equation}
T_{\psi}(s_t,a_t ) = \hat{T}_{\psi}(s_t,a_t ) + \Delta(s_t,a_t )
\end{equation}
where $\Delta (\cdot)$ represents the residual dynamics.

In our proposed knowledge-informed virtual environment model, we leverage traffic domain knowledge by employing the IDM as $\hat{T}_{\psi}$. The IDM is a well-established car-following model that captures the average dynamics of the traffic system, providing a reliable representation of the basic traffic behavior. By incorporating this domain expertise into the virtual environment, we ensure that the base model $\hat{T}_{\psi}$ reliably describes the fundamental dynamics of traffic flow. Moreover, its relatively simple formulation makes it an ideal starting point for demonstrating our framework’s capabilities. It is worth noting that the IDM can be easily extended or replaced with more sophisticated models as needed.

To further enhance the ability of the virtual environment to capture complex and uncertain dynamics, we introduce a neural network $\Delta_{\phi}$ with parameter $phi$ to learn the residual dynamics. The residual network $\Delta_{\phi}$ is designed to model the discrepancies between the predictions of IDM and the actual traffic dynamics, enabling the virtual environment to adapt to complex traffic patterns and uncertainties. Consequently, the knowledge-informed virtual environment model is formulated as:
\begin{equation}
\hat{T}_{\psi,\phi}(s_t,a_t ) = \hat{T}_{\psi}(s_t,a_t ) + \Delta_{\phi}(s_t,a_t )
\end{equation}

\subsubsection{Knowledge-driven traffic dynamics model}
IDM is a widely adopted car-following model in traffic simulation and modeling, providing valuable insights into traffic flow dynamics and the influence of different driving behaviors on traffic congestion and safety. In the IDM, drivers aim to maintain a safe and comfortable following distance from the leading vehicle while adjusting their acceleration to reach a desired velocity. The IDM takes into account factors including the relative distance and velocity to the leading vehicle, and its desired speed. The IDM equations are as follows:
\begin{equation}\label{eq5}
a^* = a_{\max} \left[ 1 - \left(\frac{v}{v_0}\right)^{\delta} - \left(\frac{s^*(v, \Delta v)}{s}\right)^2 \right]
\end{equation}
\begin{equation}\label{eq6}
s^*(v, \Delta v) = s_0 + \max \left\{ 0, v T_0 + \frac{v \times \Delta v}{2 \sqrt{a_{\max} b}} \right\}
\end{equation}
where $a^*$ is the calculated acceleration, $v$ is the current velocity of the vehicle, $a_{\max}$ the maximum acceleration, $v_0$ is the desired velocity, $\delta$ is the acceleration exponent, 
$s$ is the gap to the leading vehicle, $s^* (v,\Delta v)$ is the desired gap considering the relative velocity $\Delta v$ to the leading vehicle, $s_0$ is the minimum desired gap between the vehicle and its leading vehicle, $T_0$ denotes the safe time headway, and $b$ represents the comfortable deceleration of the vehicle.

Equation (\ref{eq5}) describes the acceleration dynamics of the vehicle, which is determined by three terms: 
(i) the first term $\left[ 1 - \left(\frac{v}{v_0}\right)^{\delta} \right]$ encourages the vehicle to reach its desired velocity $v_0$; 
(ii) the second term $\left(\frac{s^*(v, \Delta v)}{s}\right)^2$ ensures that the vehicle maintains a safe distance from the leading vehicle; 
and (iii) the maximum acceleration limit $a_{\max}$ prevents unrealistic accelerations.
Equation (\ref{eq6}) defines the desired gap $s^*(v, \Delta v)$, which is a function of the current velocity $v$ and the relative velocity $\Delta v$ to the leading vehicle. 
The desired gap consists of three components: (i) the minimum desired gap $s_0$; 
(ii) the velocity-dependent term $vT_0$, which increases the desired gap proportionally to the vehicle velocity to maintain a safe time headway; 
and (iii) the dynamic term $\frac{v \times \Delta v}{2 \sqrt{a_{\max} b}}$, which increases the desired gap when the vehicle is approaching a slower leading vehicle to ensure a safe and comfortable deceleration.

By incorporating the IDM as the knowledge-driven base model $\hat{T}_{\psi}$ in our virtual environment, we leverage the domain expertise encoded in the model to capture the average traffic dynamics. The IDM provides a reliable representation of the fundamental car-following behavior, considering factors such as desired velocity, safe distance, and comfortable acceleration and deceleration. This integration of traffic domain knowledge enhances the reliability of the virtual environment.

\subsubsection{Knowledge-informed neural network}
While the IDM model captures average behaviors in the traffic system with reasonable fidelity, it may not be sufficient for tasks that require a more accurate representation of real-world dynamics. For instance, drivers typically exhibit diverse driving styles, preferences, and situational responses. The simplified assumptions of IDM regarding uniform driving behavior may overlook these individual differences, leading to inaccuracies in real-world traffic scenarios. Furthermore, the fixed parameters in IDM, such as the desired time headway and comfortable deceleration, may not fully capture the adaptability and context-dependent nature of human driving. These limitations can result in discrepancies between the actual behaviors and the predictions of IDM.

To address these challenges, we propose a knowledge-informed neural network to learn the residual dynamics. By combining prior knowledge with a neural network, this knowledge-informed approach can improve prediction accuracy and learning efficiency. The neural network is denoted as follows:
\begin{equation}
\Delta \hat{s}^r_{t+1}, \Delta \hat{r}^r_t = \Delta_{\phi} (s_t, a_t)
\end{equation}
where $\Delta \hat{s}^r_{t+1}$ and $\Delta \hat{r}^r_t$ represent the predicted residuals of state and reward, respectively. 
The neural network $\Delta_{\phi}$ takes the current state $s_t$ and action $a_t$ as inputs and outputs the estimated residuals, which capture the discrepancies between the IDM predictions and the actual dynamics. 
The knowledge-informed neural network is trained in an end-to-end manner using the interaction data generated by the RL agent’s interactions with the environment. 
The input to the model consists of the state features. The model then directly outputs the predicted residual dynamics, which are compared with the ground truth residual dynamics to calculate the MSE (mean square error) loss for learning. 
By learning the residuals, the neural network can fine-tune the knowledge-driven IDM model to better match real-world data, thereby improving the overall fidelity of the virtual environment. The predicted residuals from the neural network are then added to the predictions of $\hat{T}_{\psi}$ to obtain the final outputs of the virtual environment model.

The integration of the knowledge-driven IDM model and the data-driven residual neural network enables the virtual environment to accurately capture both the average dynamics of the traffic system and the individual variations and adaptations of human driving behavior. The IDM serves as a base model, capturing the fundamental dynamics of traffic flow. The use of residual learning is specifically designed to address the limitations of fixed IDM parameters. The residual component, implemented as a neural network, allows our model to capture complex behaviors and dynamics that the IDM alone cannot describe. For instance, while the IDM might predict a certain acceleration based on its fixed parameters, the residual component can learn to adjust this prediction based on factors not considered by the IDM, such as individual driver characteristics. This combination allows the virtual environment to generate realistic predictions and facilitates effective learning for the RL agent. By interacting with this knowledge-informed virtual environment, the RL agent can explore and learn from a large number of simulated experiences, improving sample efficiency and reducing the need for extensive actual environment interactions.

\subsection{Driving policy learning with residual RL}
In our proposed approach, we aim to leverage existing well-established knowledge to improve the learning efficiency of RL and avoid learning from scratch. 
Given an initial policy $\pi_H: \mathcal{S} \rightarrow \mathcal{A}$ with states $s \in \mathcal{S}$ and actions $a \in \mathcal{A}$, we learn a residual policy $\pi_\theta: \mathcal{S} \rightarrow \mathcal{A}$. 
The main idea is to leverage the reliability of conventional controllers while also benefiting from the flexibility and adaptability of RL. By combining the initial policy $\pi_H$ and the residual policy $\pi_\theta$, we obtain a full policy given by:
\begin{equation}
\pi = \pi_{H} + \pi_{\theta}
\end{equation}

The initial policy $\pi_{H}$ can be any conventional controller, such as a hand-designed rule-based policy, a model-predictive controller, or any other analytical physics-based controller. These controllers often incorporate domain knowledge and are designed to provide a baseline level of performance in the given task. However, they may lack the ability to adapt to complex and changing environments or to capture intricate patterns and behaviors that are difficult to model analytically. On the other hand, the residual policy $\pi_{\theta}$ is learned using RL techniques, which enable the agent to adapt and optimize its behavior based on the interactions with the environment. The residual policy is designed to capture the discrepancies between the actions of the initial policy and the optimal actions, allowing the agent to refine and improve upon the decisions of the initial policy. 

\begin{figure*}
\centering
  \includegraphics[width=0.82\textwidth]{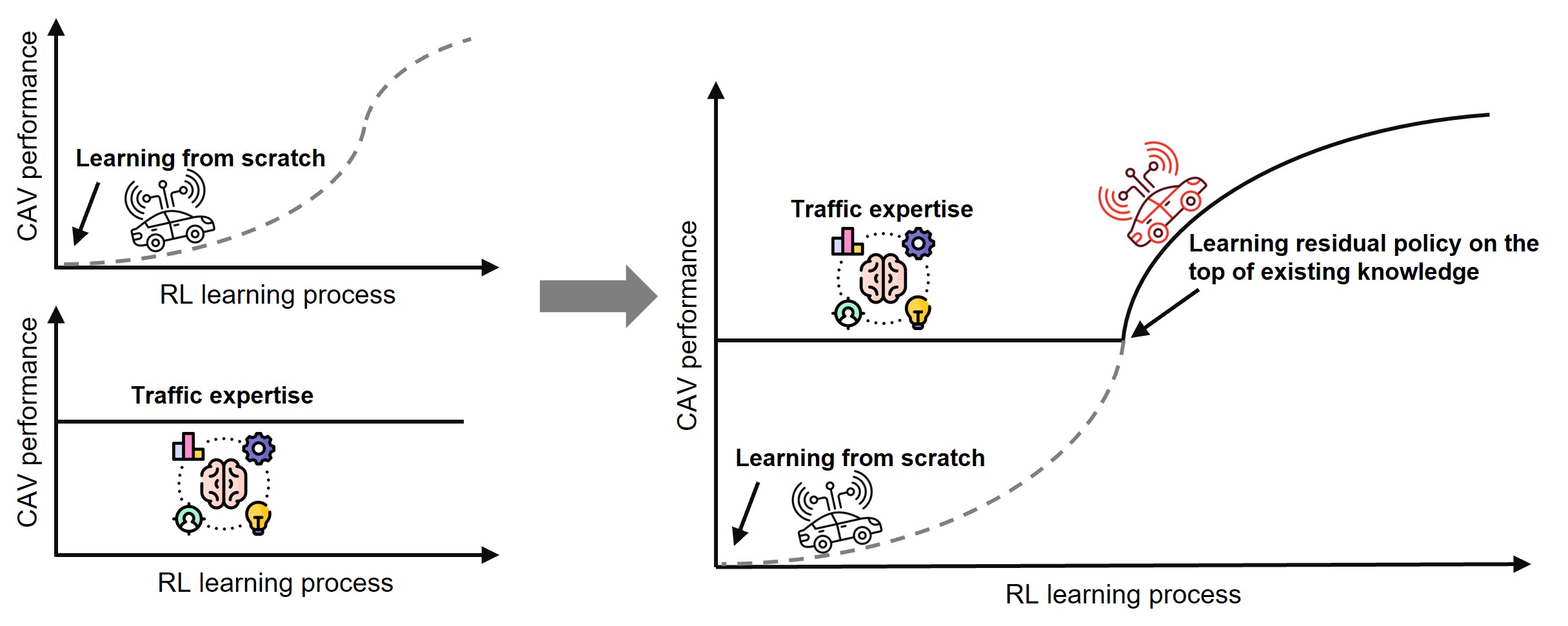}
  \caption{Comparison of learning processes between traditional RL and our proposed residual RL approach leveraging existing traffic domain knowledge.}
  \label{fig2}
\end{figure*}

As illustrated in Figure \ref{fig2}, the traditional RL starts learning from scratch, where the agent learns a policy solely based on the rewards received from the environment. 
This approach can be sample-inefficient and may require numerous interactions with the environment to converge to a satisfactory policy. 
In contrast, by incorporating an initial policy $\pi_{H}$, such as the traffic expertise in our case, the agent can start with a reasonably good baseline behavior. 
The residual policy $\pi_{\theta}$ is then learned to complement and enhance the initial policy, focusing on capturing the residual differences between the actions of the initial policy and the optimal actions. The combination of the initial policy and the residual policy offers several benefits. 
Firstly, as proven in Theorem A1, it allows the agent to take advantage of the efficiency and stability of conventional controllers, which can provide a solid foundation and a better start point for the learning process. 
Secondly, the residual policy enables the agent to adapt and optimize its behavior in a more targeted manner, focusing on learning the necessary adjustments to improve upon the initial policy. 
Theorem A2 demonstrates that our proposed policy can converge to an $\epsilon$-optimal policy, ensuring the effectiveness of the learning process. This learning paradigm aims to achieve faster convergence and improved sample efficiency compared to learning from scratch.

\subsubsection{Physics-based initial policy}
In our proposed framework, we employ a physics-based controller that incorporates traffic domain knowledge as the initial policy $\pi_H$. Specifically, we adopt the PI with saturation controller \citep{Stern2018Dissipation}, a well-established and field-validated model that has demonstrated effectiveness in managing CAV behaviors in real-world traffic scenarios. The PI with saturation controller aims to regulate the velocity of the CAV based on the estimated average speed of the leading vehicles while handling stop-and-go waves commonly encountered in traffic.

The core principle of the PI with saturation controller is to determine a command velocity $v^{\text{cmd}}$ based on a proportional-integral (PI) control mechanism. 
The average speed, denoted as $\bar{v}$, is estimated by calculating the average of the CAV velocity over a predefined time window. 
To ensure safety and effective gap management, the PI with saturation controller incorporates two key mechanisms. 
First, it defines a target velocity $v^*$  that adjusts the speed of CAV based on the current gap between the CAV and the leading vehicle, which is calculated as:
\begin{equation}
v^* = \bar{v} + v^\text{c} \times
\min\left(\max\left(\frac{s-s_{l}}{s_{u}-s_{l}},0\right),1\right)
\end{equation}
where $v^c$  is the maximum additional velocity allowed for catching up to the leading vehicle, $s$ is the current gap between the CAV and the leading vehicle, and $s_l$ and $s_u$ are the lower and upper gap thresholds, respectively. 
When the gap is below a lower threshold $s_l$, the target velocity is set to the estimated average speed $\bar{v}$, ensuring that the CAV maintains a safe distance. As the gap increases beyond $s_l$, the target velocity gradually increases up to a maximum of $\bar{v}+v^{\text{c}}$ , enabling the CAV to catch up to the leading vehicle and close the gap. 

The second mechanism involves updating the command velocity $v^\text{cmd}$  using the following rule: 
\begin{equation}
v^\text{cmd}_{t+1} = \beta_t\left(\alpha_t v^*_t + (1-\alpha_t)v^\text{lead}_t\right)
+ (1-\beta_t)v^\text{cmd}_t
\end{equation}
where $t$ denotes the time step, $v^\text{lead}$ is the velocity of the leading vehicle, and $\alpha_t$ and $\beta_t$ are weight factors determined based on the gap between the CAV and the leading vehicle. 
The weight factor $\alpha$ is calculated as: 
\begin{equation}
\alpha = \min\left(\max\left(\frac{s - \Delta x^\text{s}}{2},0\right),1\right)
\end{equation}
where $\Delta x^\text{s}$ is the safety distance. 
When the gap is below a safety distance $\Delta x^\text{s}$, $\alpha$ is set to 0, prioritizing the velocity of the leading vehicle to ensure safety. 
As the gap increases, $\alpha$ smoothly transitions to 1, allowing the CAV to adjust its speed towards the target velocity. 
The weight factor $\beta = 1 - \frac{1}{2}\alpha$ regulates the rate at which the controller responds to changing situations, with quicker adjustments taking place in more critical safety scenarios.

Extensive field experiments have validated the effectiveness of the PI with saturation controller in managing CAV behavior and enhancing traffic flow. However, this controller alone may not be sufficient to capture the full complexity and variability of real-world traffic dynamics. 
Therefore, we complement this initial policy with a residual RL agent $\pi_\theta$, which learns to adapt and optimize the behavior of the CAV based on the specific traffic conditions encountered. The combination of the physics-based controller and the residual RL agent enables the CAV to leverage prior knowledge while continuously learning and improving its decision-making capabilities. 
It is worth noting that the employed physical models require discretization or simplification to adapt to the neural networks because the reinforcement learning environment operates in a discrete manner, with fixed time steps and state transitions. However, we can directly integrate these physical models into the neural network architecture without the need for simplification. 
The advantage of this direct integration is that it preserves the accuracy and expressiveness of the physical models while allowing seamless incorporation into the residual reinforcement learning framework, enabling the agent to learn effective control policies that leverage both the prior knowledge and the data-driven adaptability of neural networks.

\subsubsection{RL-based residual policy}
The RL problem is typically formulated as a Markov Decision Process (MDP), which provides a mathematical framework for modeling decision-making. An MDP is defined by a tuple $(\mathcal{S}, \mathcal{A}, {T}, {R}, \gamma)$, where $\mathcal{S}$, $\mathcal{A}$, and ${T}$ are the state space, action space, and transition dynamics of the environment, as introduced in the previous sections. 
${R}: \mathcal{S} \times \mathcal{A} \rightarrow \mathbb{R}$ represents the reward function, and $\gamma \in [0,1]$ is the discount factor. 
The goal of the RL agent is to learn an optimal policy $\pi^*$ that maximizes the expected cumulative discounted reward over time, i.e., $\pi^* = \arg\max_{\pi} \mathbb{E}\left[\sum_{t=0}^{\infty} \gamma^t {R}(s_t, a_t) \mid \pi \right]$, where the expectation is taken over the trajectory of states and actions generated by following a policy $\pi$. 
To solve the RL problem and find the optimal policy, we often use the state-value function and the action-value function. 
The action-value function $Q^\pi(s, a) = \mathbb{E}_\pi \left[\sum_{t=0}^{\infty} \gamma^t {R}(s_t, a_t) \mid s_0 = s, a_0 = a \right]$ represents the expected cumulative discounted reward starting from state $s$, taking action $a$, and then following a policy $\pi$. 
The state-value function $V^\pi(s) = \mathbb{E}_{a_t \sim \pi(\cdot \mid s_t)} \left[Q^\pi(s_t, a_t) \mid s_0 = s \right]$ represents the expected cumulative discounted reward starting from state $s$ and following a policy $\pi$.

In our proposed framework, the residual policy network $\pi_\theta$ is represented by a fully-connected neural network with two hidden layers, each containing 64 neurons with ReLU activation functions. 
The output layer of the network has a $\tanh$ activation function to ensure that the residual actions are bounded. We employ Trust Region Policy Optimization (TRPO) \citep{Schulman2015Trust} to learn the residual policy $\pi_\theta$, which adapts to complex traffic dynamics and compensates for the limitations of the physics-based controller. 
The main idea behind TRPO is to perform constrained optimization in the policy space, ensuring that each policy update improves performance while preventing the policy from deviating too far. 
TRPO achieves this by defining a trust region that constrains the magnitude of each policy update, ensuring that the deviation between the updated and old policies is limited. Specifically, the optimization problem is formulated as follows:
\begin{equation}
\begin{aligned}
\arg\max_\theta \quad & \mathbb{E}_{\tau \sim \pi_{\theta_{\text{old}}}} \left[
\sum_{t=0}^{T} \frac{\pi_\theta (a_t \mid s_t)}{\pi_{\theta_{\text{old}}} (a_t \mid s_t)} \cdot Q^\pi (s_t, a_t) \right] \\
\text{s.t.} \quad & \mathbb{E}_{\tau \sim \pi_{\theta_{\text{old}}}} \left[ D_{KL} \left[ \pi_{\theta_{\text{old}}} (\cdot \mid s_t) \parallel \pi_\theta (\cdot \mid s_t) \right] \right] \leq \delta
\end{aligned}
\end{equation}
where $\pi_{\theta_{\text{old}}}$ and $\pi_\theta$ represent the old and updated policies, respectively. $\tau$ is the trajectory sampled from the old policy. $D_{KL}$ measures the Kullback-Leibler (KL) divergence between the old and updated policies. The constraint ensures that the policy update remains within the trust region defined by the threshold $\delta$, which helps to stabilize the learning process and prevent the policy from making drastic changes that could lead to poor performance.

The training procedure involves iteratively collecting trajectories, computing rewards, and updating the policy network. Specifically, we first initialize the policy network with random parameters. 
Then, for each training iteration, we collect a batch of trajectories by executing the current policy in the environment, where each trajectory is a sequence of states, actions, and rewards, denoted as $(s_t,a_t,r_t,s_{t+1})$. 
The rewards for each step in the trajectories are computed using the reward function defined in the environment. Next, we update the policy network using the TRPO algorithm, which involves computing the natural gradient using the conjugate gradient method and determining the step size that satisfies the KL divergence constraint through a line search. 
This process is repeated until convergence or a maximum number of iterations is reached, with convergence determined based on the improvement in the average reward over a fixed number of iterations.

\subsection{Model-based residual RL}
Despite incorporating prior knowledge and neural networks to learn the environment dynamics can improve the prediction performance, there still exists model inaccuracy in the virtual environment model due to factors such as the complexity of real-world traffic scenarios, and inherent stochasticity in human driving behavior. 
Any inaccuracies in the IDM or the neural network components could significantly impact the performance of the RL agent. This is because the agent learns and optimizes its policy based on the interactions with both the virtual and actual environments. If the virtual environment does not accurately reflect real-world traffic dynamics, the learned policy may not transfer well to actual traffic scenarios. 
Inaccuracies could lead to suboptimal decision-making, reduced safety, or inefficient traffic flow when the agent is deployed in the actual environment. Therefore, we aim to design a feasible model-based RL approach that can effectively handle and mitigate the impact of these inaccuracies, ensuring that the policy performance improvement observed in virtual environment can be transferred to the actual environment.

As the number of rollout steps $k$ in the virtual environment model increases, a trade-off emerges between the growing cumulative error and the potential for higher policy performance improvement due to the increased virtual interaction data. 
On one hand, longer rollouts provide more diverse and informative experiences for the RL agent to learn from, leading to a more optimal policy. On the other hand, the accumulation of model inaccuracies over extended rollouts can negatively impact the reliability and effectiveness of the learned policy. 
Consequently, it is crucial to balance these opposing factors to ensure that the benefits of longer rollouts outweigh the detrimental effects of the increased model error. To guarantee policy improvement in the actual environment, the policy performance must exceed a certain threshold $C$ in the virtual environment \citep{Janner2019When}, which is determined by the model inaccuracy, the rollout length, and the discrepancy between the updated and previous policies. Based on these, we can derive the following relationship:
\begin{equation}
\eta(\pi) \geq \hat{\eta}(\pi) - C(\epsilon_\pi, \epsilon_m, k)
\end{equation}
\begin{equation}
C(\epsilon_\pi, \epsilon_m, k) = 2R_{\text{max}} \left[\frac{\gamma^{k+1} \epsilon_\pi}{(1-\gamma)^2} + \frac{\gamma^k \epsilon_\pi}{1-\gamma} + \frac{k \epsilon_m}{1-\gamma}\right]
\end{equation}
where $\eta(\pi)$ and $\hat{\eta}(\pi)$ represent the cumulative reward in the actual and virtual environment, $k$ denotes the virtual model rollout length, 
$\epsilon_\pi$ denotes the bounded total-variation distance due to the policy update, defined as $\epsilon_\pi = \max_s D_{\text{TV}} \left[\pi_\theta \parallel \pi_{\theta_{\text{old}}} \right]$, 
and $\epsilon_m$ is the bounded model inaccuracy. 
This relationship provides insights for how to utilize the virtual model for policy update, i.e., we should minimize $C$ throughout the learning process to maximize the lower bound of $\eta(\pi)$. 
Although analytically solving for the minimum of $C$ is challenging, we observe that $C$ increases monotonically with respect to the model inaccuracy $\epsilon_m$ and can grow to infinity as the rollout length $k$ increases. 
Consequently, it is advisable to prevent the excessive use of the virtual environment model, and reduce the rollout length $k$ when faced with high model inaccuracy. Based on this, we propose a dynamic rollout length as follows:
\begin{equation}\label{eq15}
k^* = \min \left(k_{\text{max}}, \left\lfloor \frac{\kappa}{\epsilon_m} \right\rfloor \right)
\end{equation}
where $k_{\max}$ is a predefined upper bound for rollout step to avoid model overusing, 
$\kappa$ is a hyperparameter that controls the sensitivity of the rollout length to the model inaccuracy. 
When the model inaccuracy $\epsilon_m$ is low, the rollout length $k^*$ is allowed to increase, enabling the RL agent to benefit from longer rollouts and potentially achieve higher policy performance improvement. 
Conversely, when the model inaccuracy $\epsilon_m$ is high, the rollout length $k$ is reduced to mitigate the negative impact of accumulated errors on the policy improvement process. 
Theorem A3 elucidates the effectiveness of our virtual environment model in policy optimization, proving that policies learned in the virtual environment can effectively transfer to the actual environment.

Combining the above proposed components, the overall procedure of our knowledge-informed residual reinforcement learning framework is summarized in Algorithm \ref{algo1}.

\begin{algorithm}
\caption{Knowledge-informed model-based residual RL}
\label{algo1}
\begin{algorithmic}[1]
\State Initialize residual policy network $\pi_\theta$, value network $q_\varphi$, virtual environment model $\hat{T}_{\psi,\phi}$, and physics-based policy $\pi_H$;
\State Initialize experience buffers $\mathcal{D}_a \gets \emptyset$ for storing actual interaction data and $\mathcal{D}_v \gets \emptyset$ for virtual interaction data;
\Repeat
    \State Collect transitions from the actual environment using $\pi_\theta$ and $\pi_H$, and add to $\mathcal{D}_a$;
    \State Sample a transition batch from $\mathcal{D}_a$ to train the virtual model $\hat{T}_{\psi,\phi}$;
    \State Get rollout length $k^*$ using Equation (\ref{eq15});
    \State Collect virtual transitions using $\hat{T}_{\psi,\phi}$, $\pi_\theta$ and $\pi_H$, and add to $\mathcal{D}_v$;
    \State Sample transition batches from $\mathcal{D}_a$ and $\mathcal{D}_v$ to update $\pi_\theta$ and $q_\varphi$;
\Until the policy performs well in the actual environment.
\end{algorithmic}
\end{algorithm}

\section{Experimental results and analysis}\label{sec4}
\subsection{Experimental setup}
To validate the proposed method, we utilize an open-source traffic simulator SUMO \citep{Lopez2018Microscopic} in conjunction with an RL framework Flow \citep{Wu2022Flow} to construct the RL environment and control the CAV agent. 
SUMO provides a comprehensive platform for simulating a wide range of traffic scenarios in both urban and highway networks, allowing for realistic modeling of traffic dynamics and interactions between vehicles. 
Flow is built on top of the SUMO simulator and offers a user-friendly and extensible interface specifically designed for researchers to develop and evaluate advanced traffic control strategies using RL techniques. 
By leveraging the capabilities of SUMO and Flow, we create a flexible experimental setup that enables the effective validation of our knowledge-informed residual model-based RL approach in various traffic settings. 
All the parameters utilized in the experimental evaluation are summarized in \ref{appdb}.

\subsubsection{Experimental scenarios}
\begin{figure*}
\centering
\subfigure[Ring]{
\includegraphics[height=3.91cm]{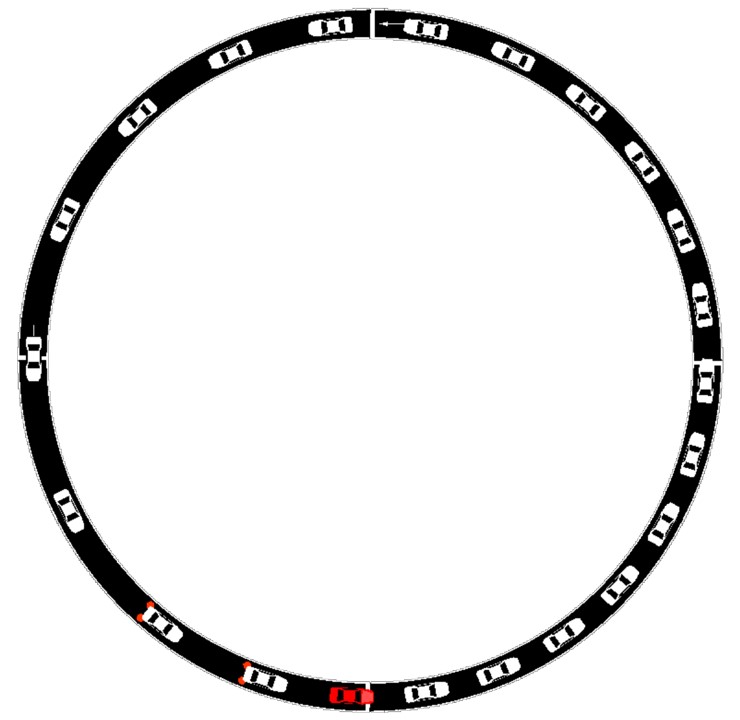}
}
\quad
\subfigure[Figure eight]{
\includegraphics[height=3.91cm]{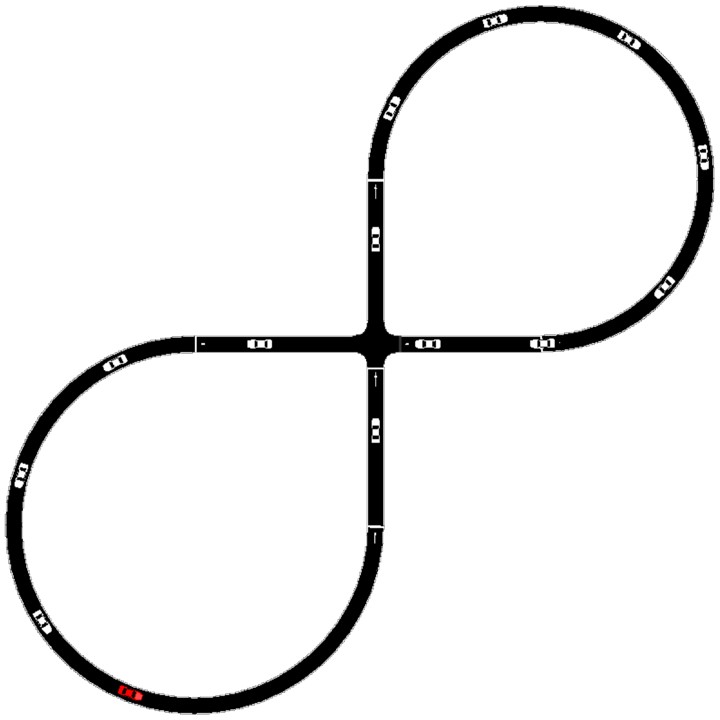}
}
\quad
\subfigure[Merge]{
\includegraphics[height=3.91cm]{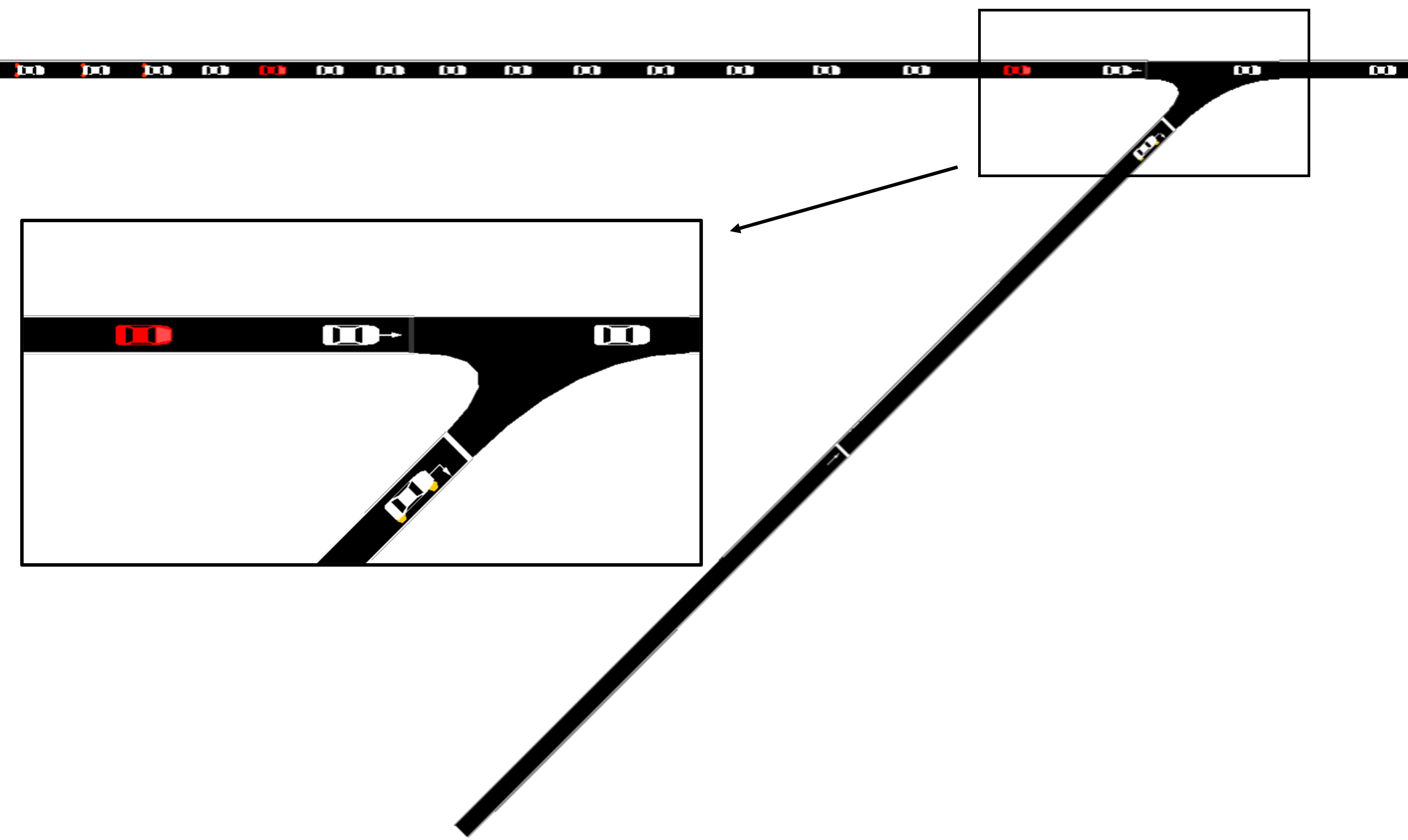}
}
\caption{Task configuration of experimental scenarios in SUMO. The red vehicle denotes CAV, and white vehicles are HDVs.}
\label{fig3}
\end{figure*}

We aim to validate that the proposed approach can mitigate traffic disturbances and smooth traffic flow in mixed traffic scenarios. As illustrated in Figure \ref{fig3}, we mainly study three scenarios: a ring road, a figure eight road, and a merge road. Each scenario includes both CAV agents and human-driven vehicles (HDVs) to form mixed traffic. The CAV agents are controlled by the RL algorithm, while HDVs are governed by the IDM. 
The IDM parameters used for human drivers are set to typical values for highway driving. To introduce stochasticity and account for real-world uncertainties, we augment the IDM by adding Gaussian noise with a mean of 0 and a standard deviation of 0.2 to the vehicle acceleration \citep{Wu2022Flow}. This noise injection allows us to simulate the variability and randomness present in actual traffic scenarios, enhancing the robustness and realism of our simulations.

\textbf{Ring Road:} The ring road scenario is designed to simulate traffic flow dynamics on a single-lane circular roadway. The total length of the ring road varies between 220 and 270 meters, providing a compact yet representative environment for studying traffic congestion. The ring road length varies across different episodes, introducing diverse spatial dynamics. 
In this scenario, we follow the setting of existing work \citep{Stern2018Dissipation}, considering a fleet of 22 vehicles, where one vehicle is designated as a CAV agent, and the remaining 21 vehicles are modeled as HDVs adhering to the IDM. 
The high vehicle density on the ring road, coupled with the limited perception capabilities of the IDM-driven vehicles, frequently results in the formation of stop-and-go waves. These waves are characterized by alternating periods of deceleration and acceleration, resulting in reduced traffic flow and increased travel times. 
The primary goal of deploying a CAV agent in this scenario is to explore the potential of our proposed approach in attenuating these stop-and-go waves and promoting a smoother, more efficient traffic flow. By leveraging the advanced control strategies and the ability to anticipate and respond to traffic conditions, the CAV agent aims to regulate the overall traffic dynamics and mitigate the negative impacts of traffic disturbances on the ring road. 

\textbf{Figure Eight:} The figure eight network presents a more complex and challenging scenario compared to the ring road. It consists of two circular loops with equal radii, ranging from 32 to 35 meters, connected at opposite ends to form a closed network resembling an unsignalized intersection. This network accommodates a total of 14 vehicles, with one vehicle designated as a CAV agent and the remaining 13 vehicles modeled as HDVs following the IDM. 
The unsignalized intersection at the center of the figure eight network serves as a potential bottleneck, as vehicles approaching from different directions must navigate through the shared space while adhering to right-of-way rules. This interaction can lead to the formation of queues and the emergence of congestion patterns, particularly when vehicles arrive at the intersection simultaneously. 
By introducing a CAV agent in this scenario, we aim to investigate how our proposed approach can alleviate congestion, optimize traffic flow, and improve the overall efficiency of the network.

\textbf{Merge:} The merge network represents a common highway scenario where an on-ramp is connected to the main highway, allowing vehicles to enter and merge with the main traffic flow. This network comprises a 700-meter single-lane highway and a 100-meter on-ramp. The inflow rates for the highway and the ramp are set to 2000 and 100 veh/hr, respectively. 
The mix of traffic in this scenario consists of 10\% CAVs on the main highway, while the remaining vehicles, both on the highway and the on-ramp, are modeled as HDVs. These CAVs allow us to investigate how a small proportion of intelligently controlled vehicles can influence the broader traffic dynamics in a more complex scenario. Moreover, while the penetration rate of CAVs is fixed, their positions within the traffic flow are randomized for each experiment. 
The merge network aims to capture the influence of merging vehicles on the main traffic flow, as the perturbations caused by the entering vehicles can trigger the emergence and spread of stop-and-go waves upstream of the merge point. These disturbances can greatly reduce the overall throughput and efficiency of the highway network. By employing CAV agents in this scenario, we investigate the potential of our proposed approach to alleviate the negative effects of merging traffic and enhance the overall traffic flow dynamics. 

These scenarios encapsulate essential elements of urban traffic complexity, including high traffic density, multiple conflict points, and the need for cooperative behavior among vehicles. They allow us to rigorously test our algorithm’s ability to handle key challenges such as congestion mitigation, intersection management, and smooth traffic flow coordination. Moreover, the use of these controlled scenarios enables us to conduct detailed analyses of our algorithm’s performance, isolating the effects of specific traffic patterns and vehicle interactions. This approach allows for a more precise evaluation of our method’s strengths and limitations, which is crucial in the development stage of new algorithms.

\subsubsection{RL settings}
\textbf{State Space:} In all the studied scenarios, each CAV agent can perceive its surrounding environment, which includes the states of the leading and following vehicles. Specifically, the state space consists of the velocity of the CAV, and the relative velocity to its leading and following vehicles, as well as the gaps between the CAV and its neighboring vehicles. This local state representation allows the CAV agent to make informed decisions based on its immediate traffic context.

\textbf{Action Space:} The action space for each CAV agent is defined as the acceleration, with a range of $\pm$1$\,\mathrm{m/s^2}$. This range is designed to maintain passenger comfort, which is a crucial factor in the widespread adoption and acceptance of automated driving systems. 
Furthermore, this range aligns with the acceleration capabilities of many production vehicles, ensuring that our approach remains practical for real-world implementation. By choosing an appropriate acceleration value at each time step, the CAV agent can control its speed and adjust its position relative to the surrounding vehicles. This continuous action space enables the CAV agent to smoothly and effectively control its motion in response to the prevailing traffic conditions.

\textbf{Reward Function:} The reward is designed to align the behavior of the RL agent with the primary objectives of improving mobility, smoothing traffic flow, and ensuring safe operations. As shown in Equation~(\ref{eq16}), the reward function is composed of three components:
\begin{equation}\label{eq16}
r = \alpha \cdot \max\left(v_{\text{des}} - \frac{1}{n} \sum_{i=1}^{n} v_i, 0\right) - \beta \cdot \sum_{j=1}^{m} \max\left(h_{\text{max}} - h_j, 0\right) - \gamma \cdot \sum_{k=1}^{m} |a_k|
\end{equation}
where $v_{\text{des}}$ is the desired velocity, and $h_{\text{max}}$ is a predefined headway threshold. The first term incentivizes the average velocity of all vehicles to approach the desired velocity and penalizes the excessively high speeds caused by vehicle collisions. The second term introduces a penalty for congested traffic, which is only applied when the headways of CAVs fall below the threshold $h_{\text{max}}$. The third term suppresses excessive or frequent acceleration and deceleration by penalizing the absolute value of the acceleration of CAV to improve driving smoothness and passenger comfort. The weighting factors $\alpha$, $\beta$, and $\gamma$ are used to balance the different terms in the reward function.

\subsection{Experimental evaluation}
\subsubsection{Comparative evaluation}
To demonstrate the effectiveness and superiority of our proposed knowledge-informed residual reinforcement learning approach, we conduct a comprehensive comparative evaluation against well-established baseline models, including Soft Actor-Critic (SAC) \citep{Haarnoja2018Soft}, Proximal Policy Optimization (PPO) \citep{Schulman2017Proximal}, and TRPO \citep{Schulman2015Trust}. These baseline models are widely recognized in the reinforcement learning community and have been successfully applied to various control tasks, making them suitable benchmarks for assessing the performance of our approach.

The primary objectives of our proposed approach are to improve traffic mobility, reduce traffic oscillations, and ensure comfortable operations in mixed traffic environments. To align our evaluation metrics with these objectives, we consider three key performance indicators: the reward value, the average velocity of all vehicles, and the standard deviation of velocities. The reward function, as defined in Equation (\ref{eq16}), provides a comprehensive measure of the performance. 
The average velocity of all vehicles serves as an indicator of overall traffic mobility, with higher values suggesting more efficient traffic flow. The standard deviation of velocities, on the other hand, quantifies the degree of traffic oscillations, with lower values indicating more stable and smoother traffic patterns. Safety is also a critical factor in real-world applications of CAVs. 
However, we observed that even in scenarios where baseline methods performed relatively poorly in terms of traffic efficiency, the collision rates remained low. Since the RL agents’ acceleration range is constrained to [-1, 1]$\,\mathrm{m/s^2}$, they tend to maintain lower speeds when RL agents perform suboptimally. 
As a result, while they may contribute to traffic congestion, they rarely cause collisions. Given that the collision rates did not provide significant discriminative information for our comparative analysis, we will not include detailed safety metrics. To ensure the reliability and robustness of our experimental results, we repeat each experiment three times using different random seeds and report the average performance across these runs. The standard deviation of the performance metrics obtained from these three independent runs is visualized by the shaded areas in figures. 
This approach allows us to assess the consistency of our proposed method across multiple runs and account for the inherent stochasticity in the reinforcement learning process. By reporting the average performance along with the standard deviation, we provide a comprehensive evaluation of the effectiveness and stability of our approach.

\begin{figure*}[!t]
\centering
\subfigure[Ring scenario]{
\includegraphics[width=0.298\textwidth,height=3.82cm]{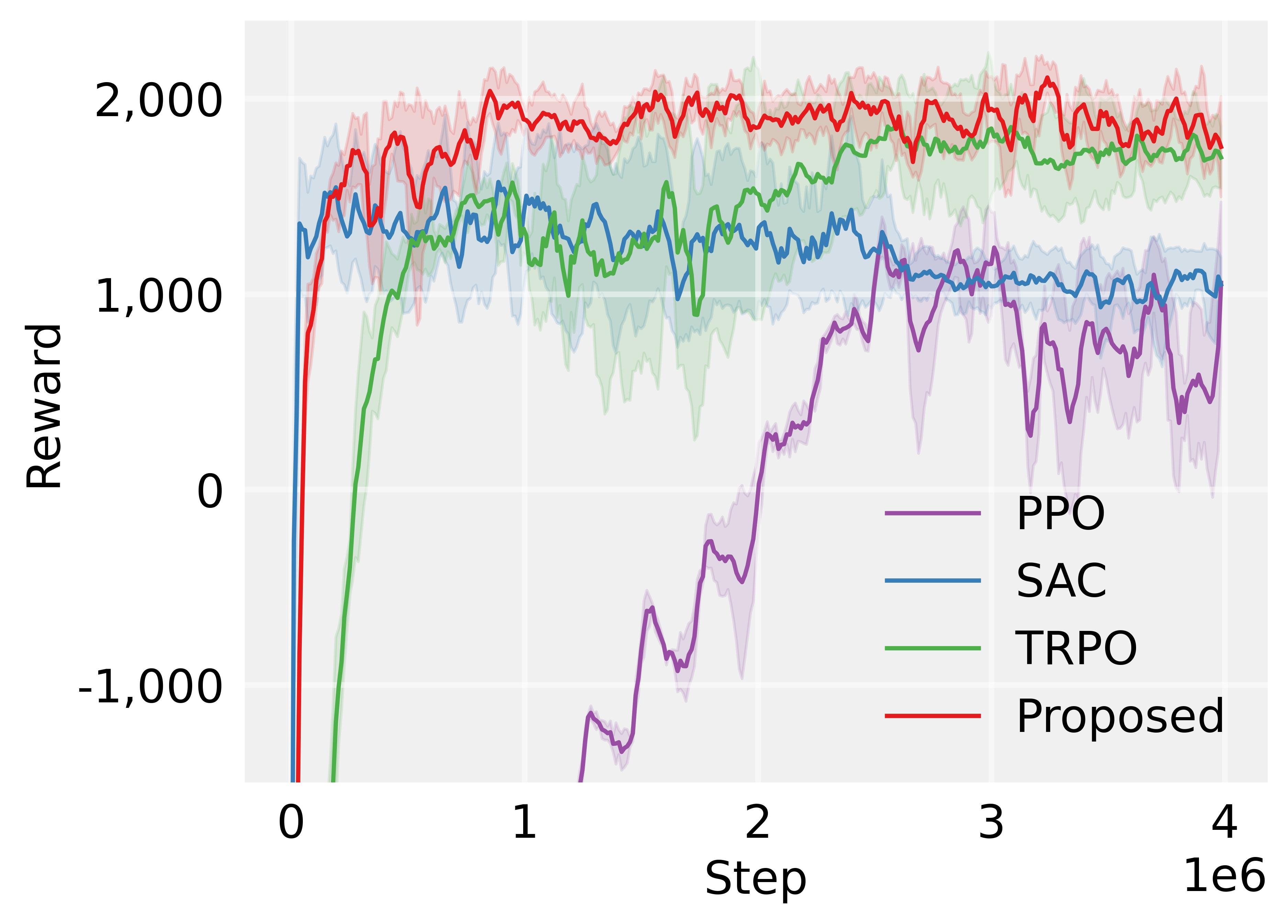}
}
\quad
\subfigure[Ring scenario]{
\includegraphics[width=0.298\textwidth,height=3.82cm]{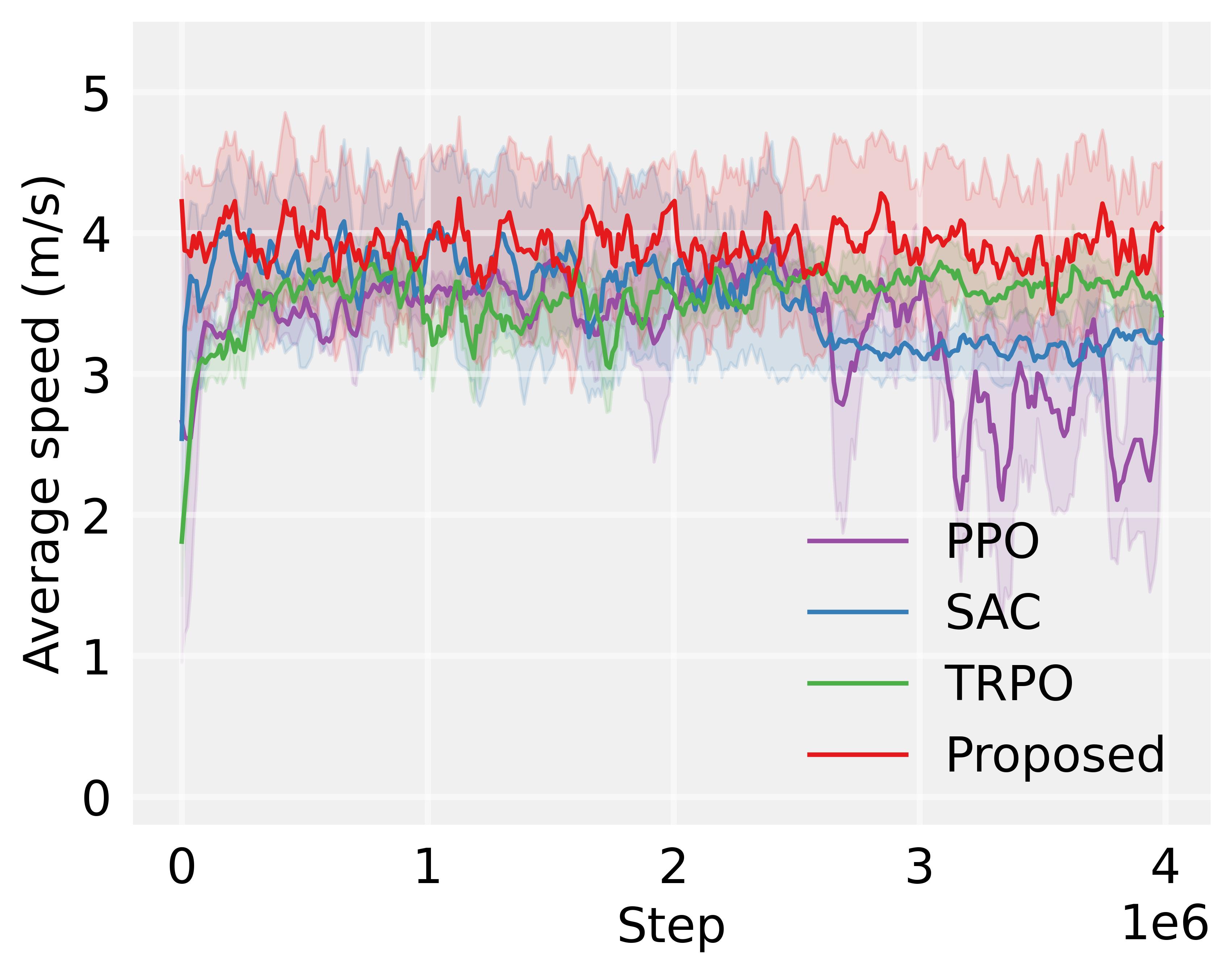}
}
\quad
\subfigure[Ring scenario]{
\includegraphics[width=0.298\textwidth,height=3.82cm]{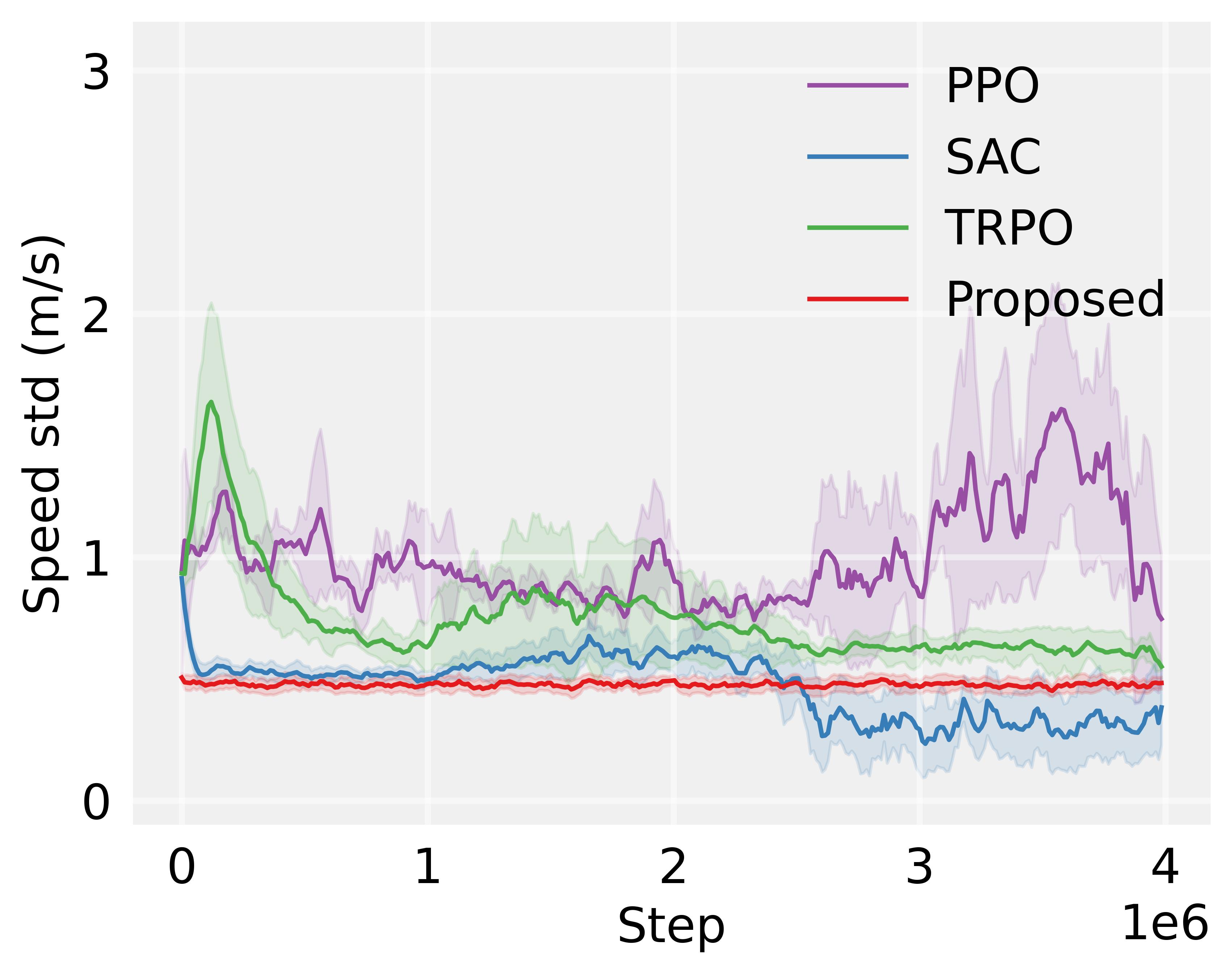}
}
\quad
\subfigure[Figure eight scenario]{
\includegraphics[width=0.298\textwidth,height=3.82cm]{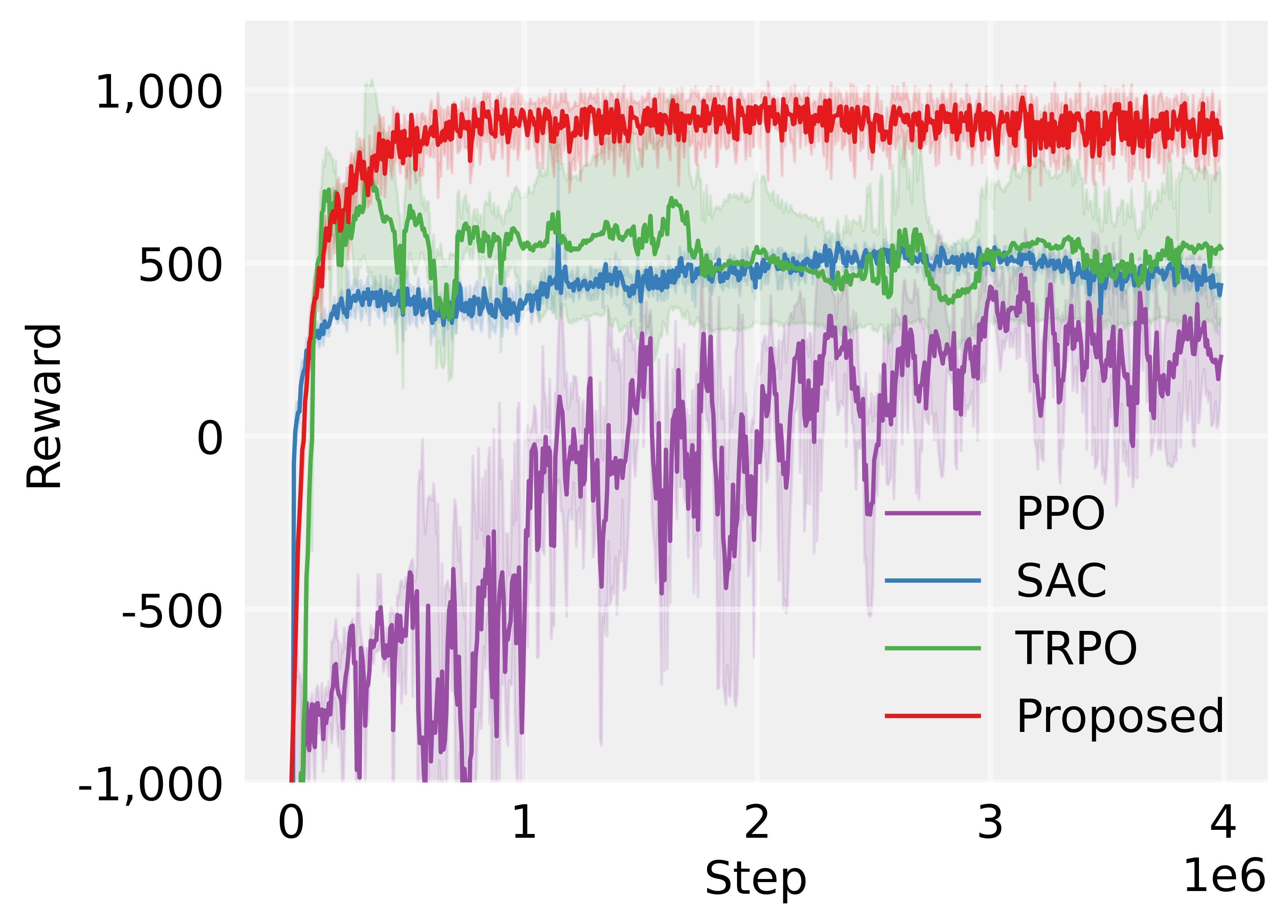}
}
\quad
\subfigure[Figure eight scenario]{
\includegraphics[width=0.298\textwidth,height=3.82cm]{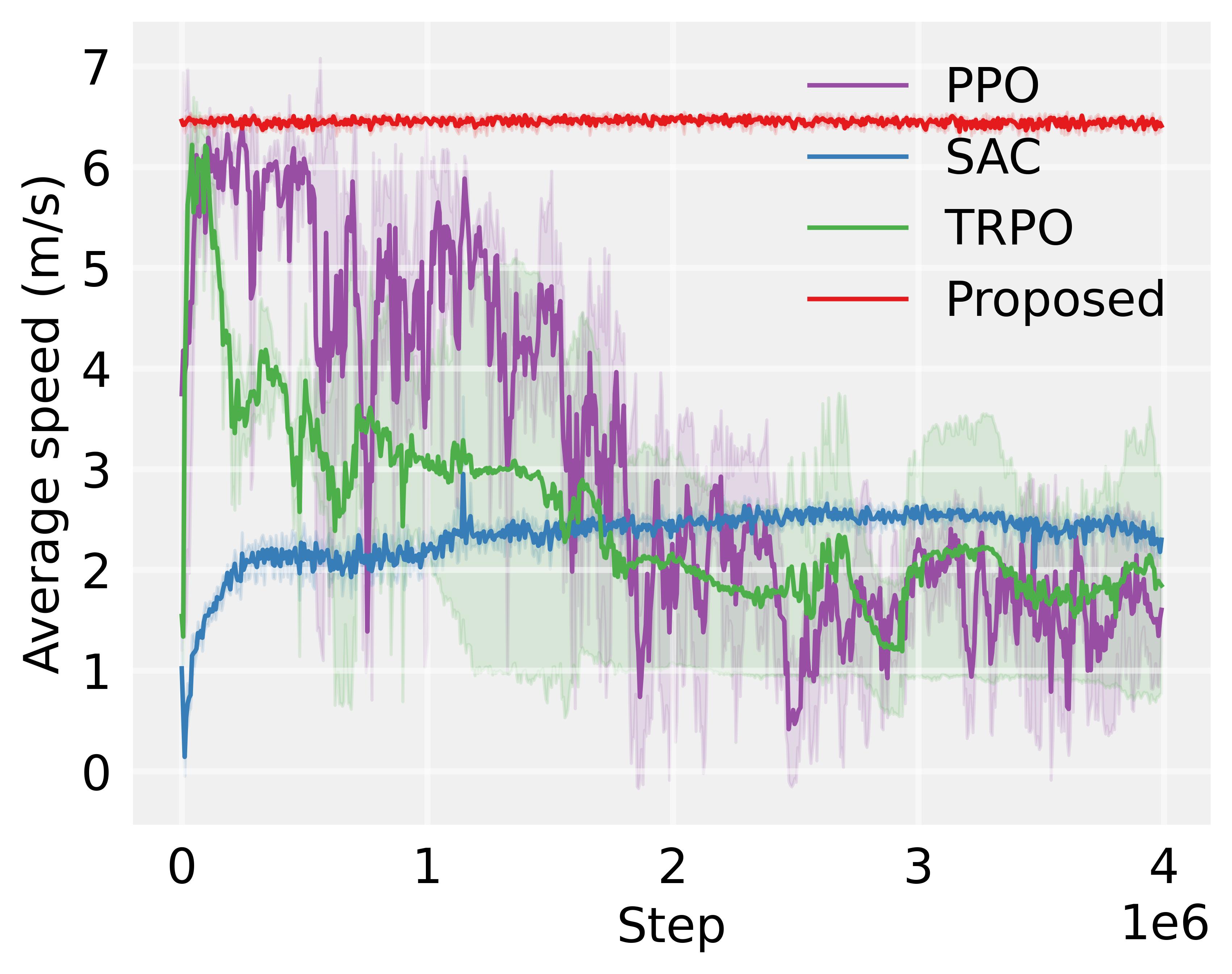}
}
\quad
\subfigure[Figure eight scenario]{
\includegraphics[width=0.298\textwidth,height=3.82cm]{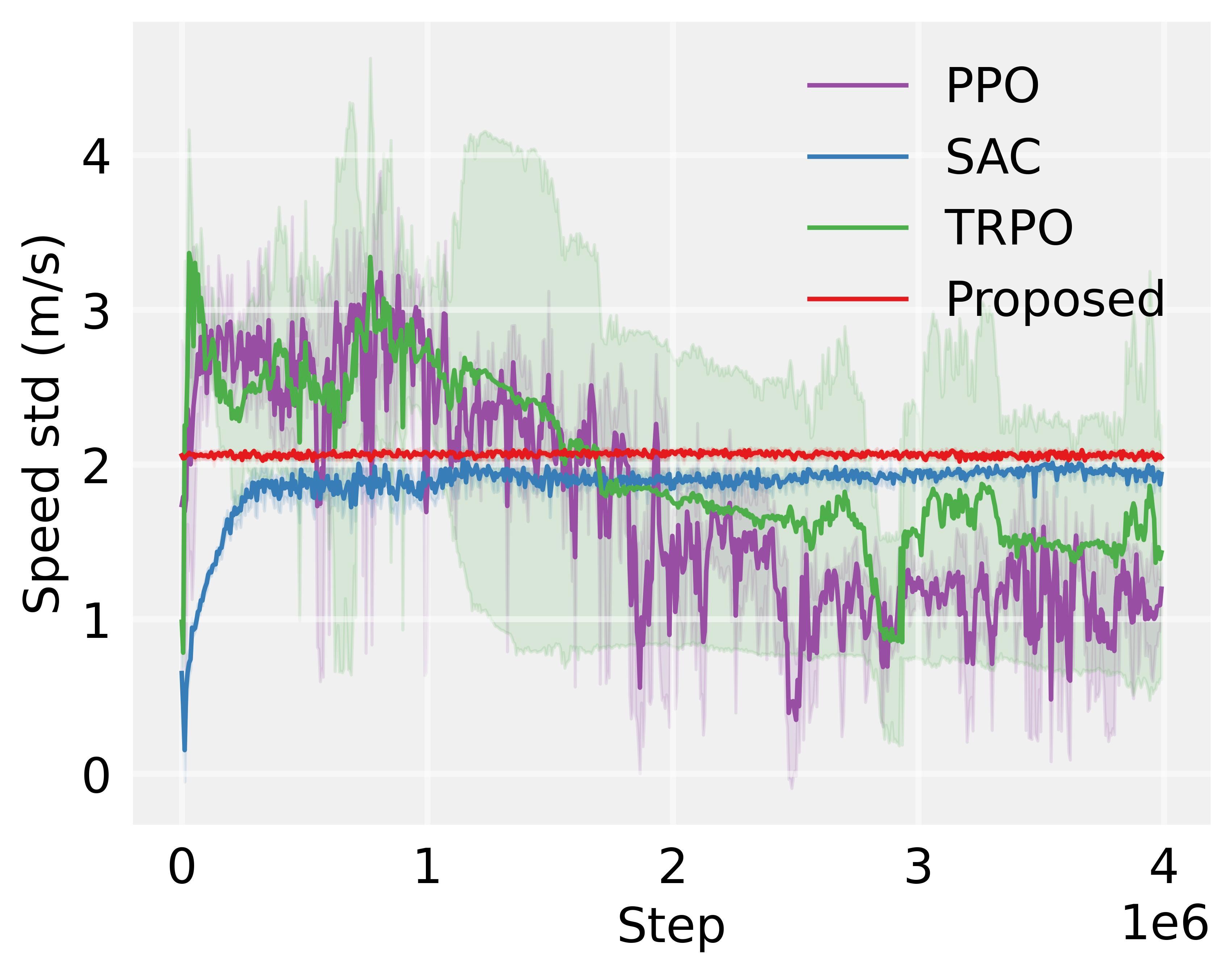}
}
\quad
\subfigure[Merge scenario]{
\includegraphics[width=0.298\textwidth,height=3.82cm]{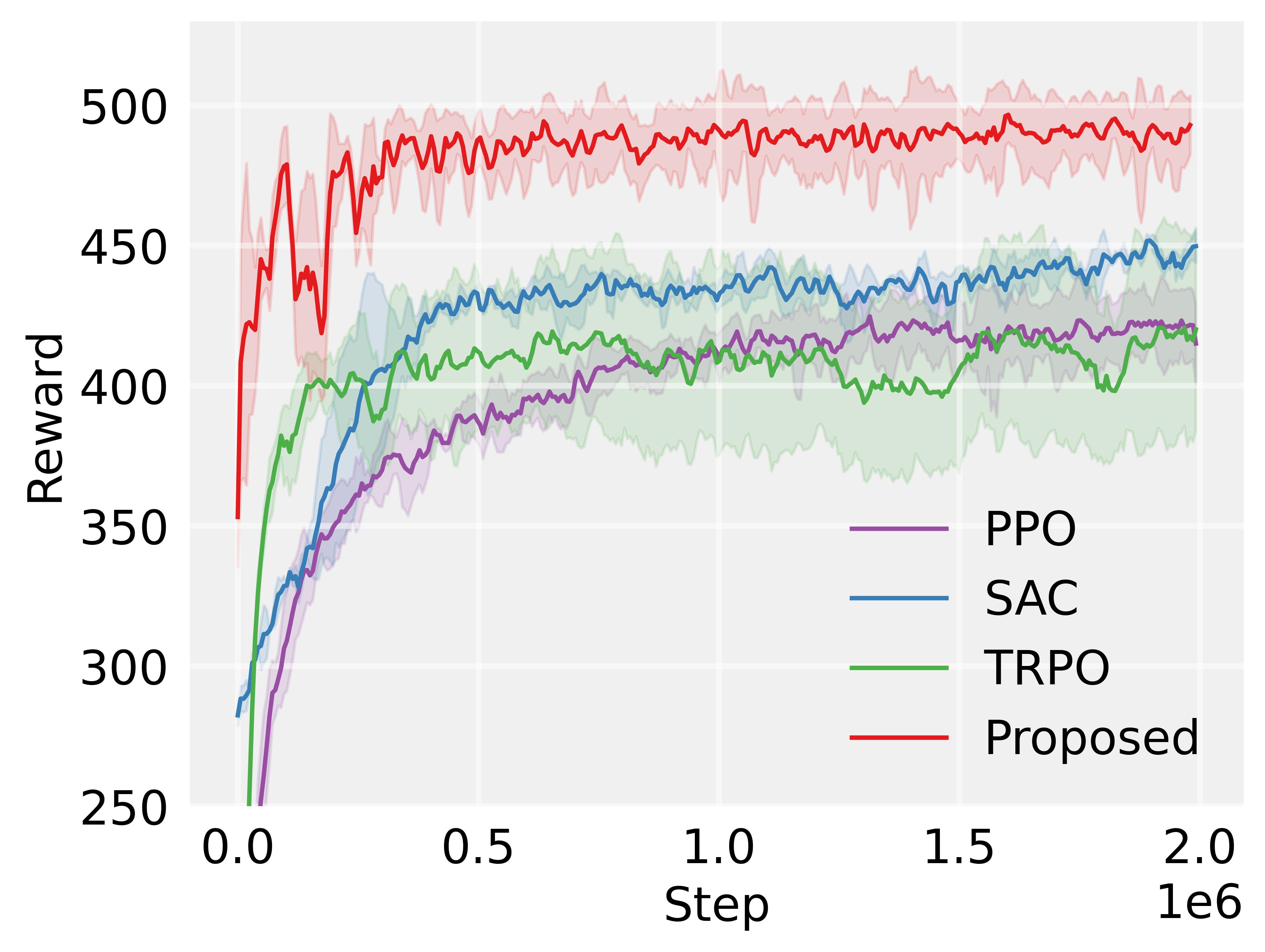}
}
\quad
\subfigure[Merge scenario]{
\includegraphics[width=0.298\textwidth,height=3.82cm]{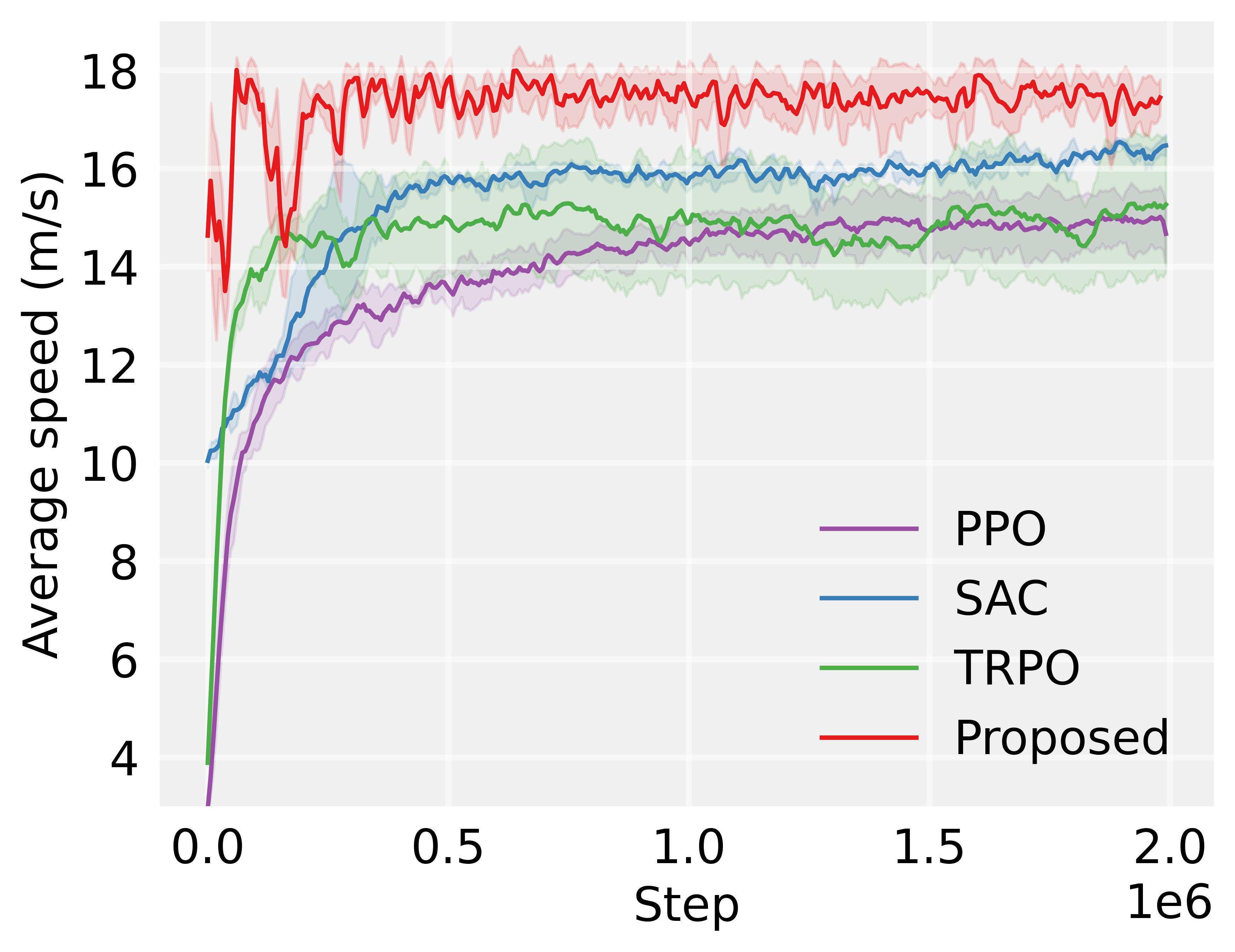}
}
\quad
\subfigure[Merge scenario]{
\includegraphics[width=0.298\textwidth,height=3.82cm]{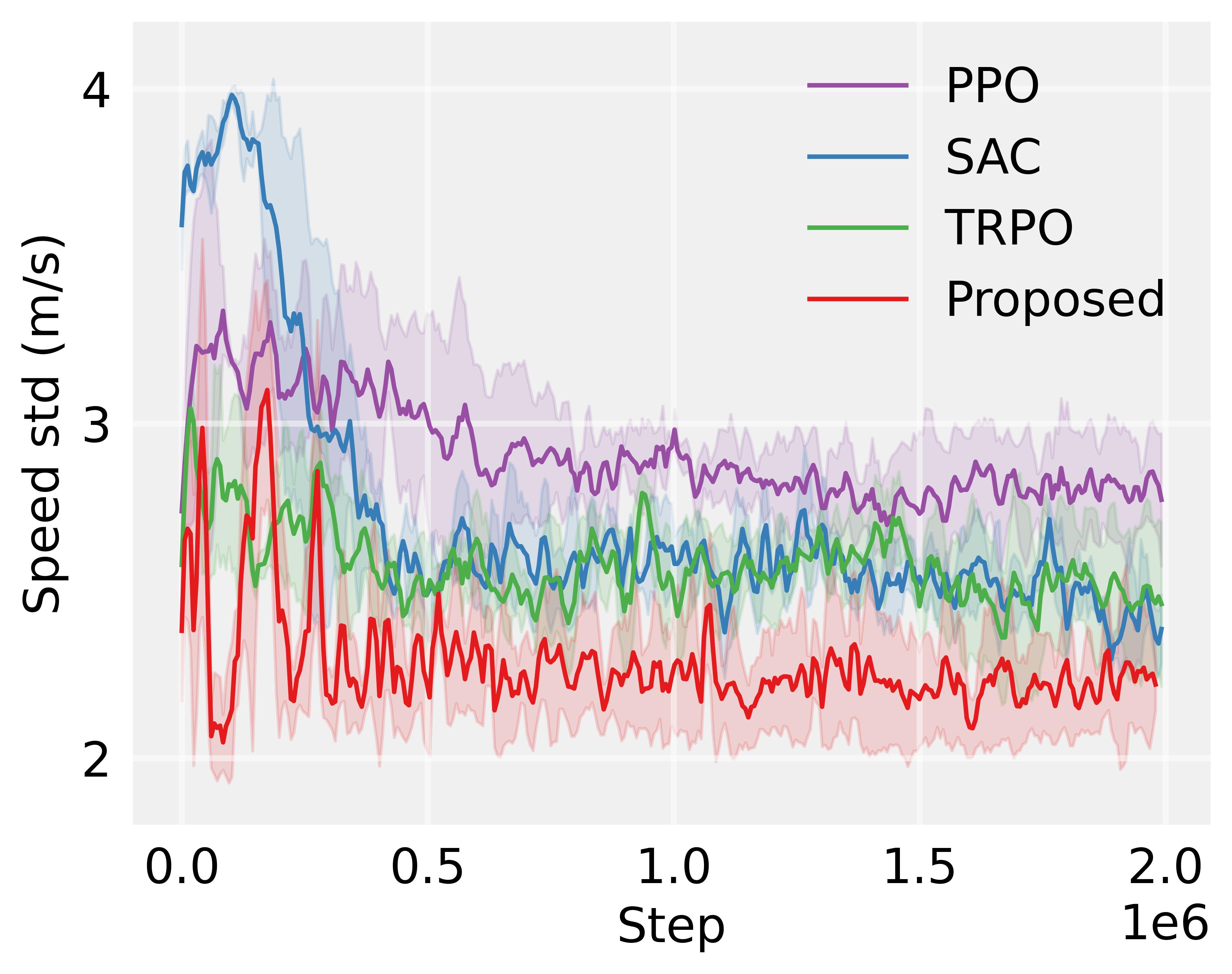}
}
\caption{Performance comparison of the proposed approach with baseline algorithms in three scenarios.}
\label{fig4}
\end{figure*}

Table \ref{tab1} and Figure \ref{fig4} demonstrate that our proposed method exhibits the best performance and superior sample efficiency compared to baseline methods across all three scenarios. This showcases the effectiveness of our approach in learning efficient control policies for mixed traffic environments. Comparing the reward values and curves, our proposed algorithm achieves higher reward values and faster convergence than the other three baseline algorithms in all scenarios. This indicates that our algorithm has higher learning efficiency and can learn better strategies within a limited number of interaction steps. Specifically, our method achieves higher rewards in the first few rounds of learning compared to the baseline models. This can be attributed to the integration of traffic expert knowledge as prior information, which allows the agent to avoid starting exploration from scratch, thereby significantly improving learning efficiency. By leveraging domain expertise, our approach enables the agent to make better decisions from the beginning, resulting in a faster rise in reward. In contrast, PPO exhibits a slower increase in reward and the final reward is 42\% lower than our proposed approach. Moreover, it can be observed that although the reward of PPO keeps increasing in the whole stage of training, its average velocity decreases in the later stages, particularly in the ring and figure eight scenarios. This indicates that PPO fails to learn an effective policy. Similarly, while the SAC algorithm shows an increase in reward and average speed during the training process in the figure eight environment, we can find that the standard deviation of velocity also increases. This suggests that the SAC algorithm does not successfully learn how to smooth traffic flow. On the other hand, our method maintains a faster rise in reward throughout the training process without experiencing performance degradation. This can be attributed to the high sample efficiency of model-based RL and the stability provided by the physics-based initial policy.

\begin{table}[!t]
\centering
\caption{Performance comparison of the proposed approach with baseline algorithms (SAC, PPO, TRPO) in three scenarios at the last training checkpoint.}
\label{tab1}
\small
\begin{tabular}{lccccccccc}
\toprule
\multirow{2}{*}{Approach} & \multicolumn{3}{c}{Ring} & \multicolumn{3}{c}{Figure eight} & \multicolumn{3}{c}{Merge} \\
\cmidrule(r){2-10}
 & {Reward} & {Avg speed} & {Speed std} & {Reward} & {Avg speed} & {Speed std} & {Reward} & {Avg speed} & {Speed std} \\
\midrule
TRPO      & 1,699.15 & 3.42 & 0.55 & 542.42 & 1.84 & 1.43 & 420.07 & 15.26 & 2.46 \\
PPO       & 1,057.14 & 3.43 & 0.75 & 228.86 & 1.60 & 1.20 & 415.02 & 14.67 & 2.77 \\
SAC       & 1,048.97 & 3.24 & 0.38 & 435.72 & 2.30 & 1.94 & 449.72 & 16.47 & 2.38 \\
Proposed  & 1,753.66 & 4.04 & 0.48 & 862.81 & 6.41 & 2.04 & 492.93 & 17.44 & 2.22 \\
\bottomrule
\end{tabular}
\end{table}

The superior performance of our approach is further highlighted by the smoother and more stable velocity curves, as shown in Figure \ref{fig4}(b), (e), and (h). The average velocity of vehicles controlled by our method is consistently higher than those of the baseline models, indicating improved traffic flow and reduced congestion. Additionally, the lower velocity standard deviation (Figure \ref{fig4}(c), (f), and (i)) demonstrates the ability of our approach to mitigate stop-and-go waves and maintain a more uniform traffic flow. The combination of model-based RL and a physics-based initial policy derived from traffic expert knowledge proves to be a powerful framework for learning effective control strategies in mixed traffic environments. By leveraging the strengths of both data-driven and knowledge-based approaches, our method achieves a faster convergence speed, higher sample efficiency, and more stable performance compared to the baseline models.

\begin{figure}[!t]
\centering
\subfigure[Ring scenario]{
\includegraphics[width=0.441\textwidth,height=4.21cm]{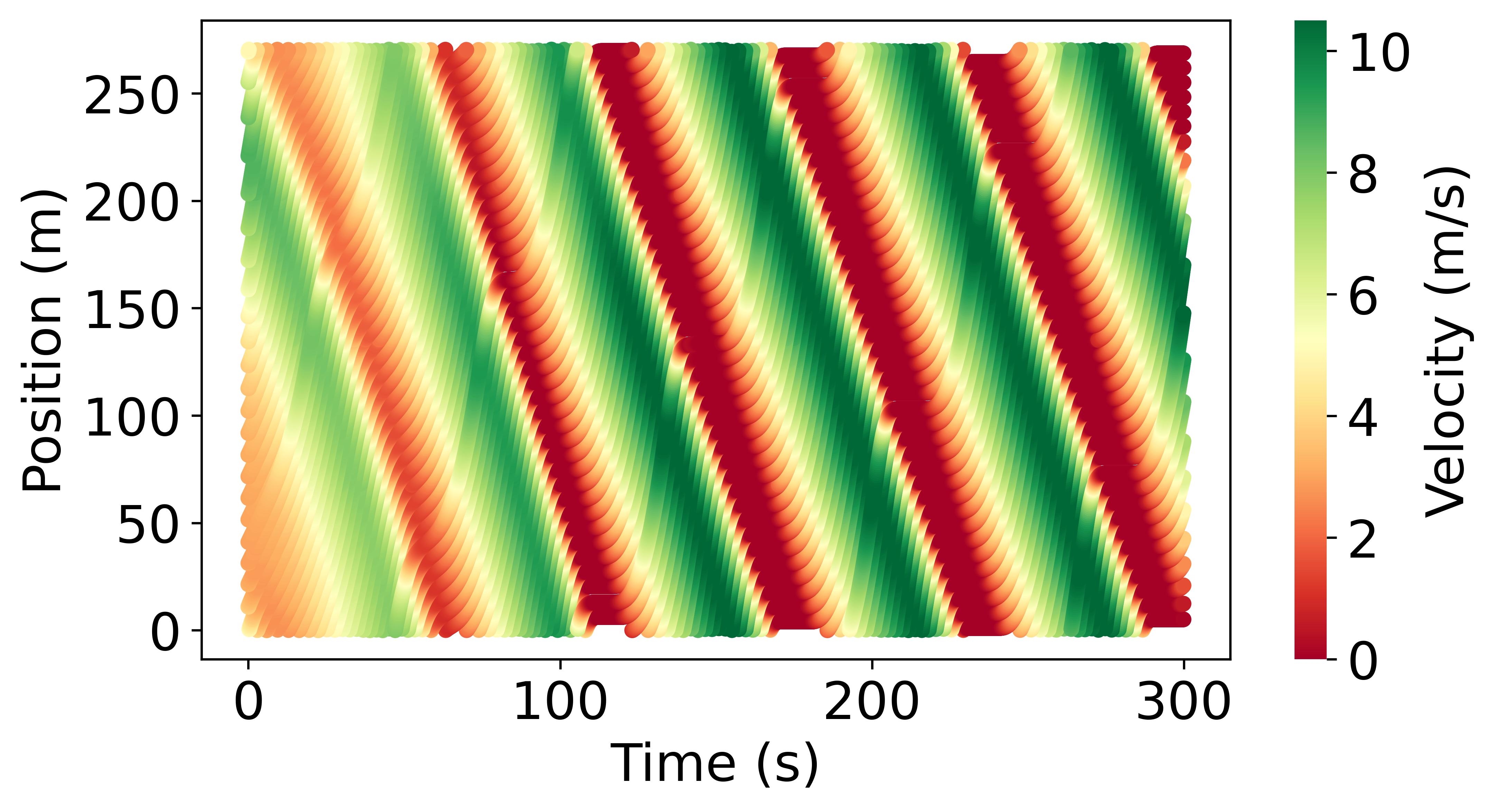}
}
\quad
\subfigure[Ring scenario]{
\includegraphics[width=0.441\textwidth,height=4.21cm]{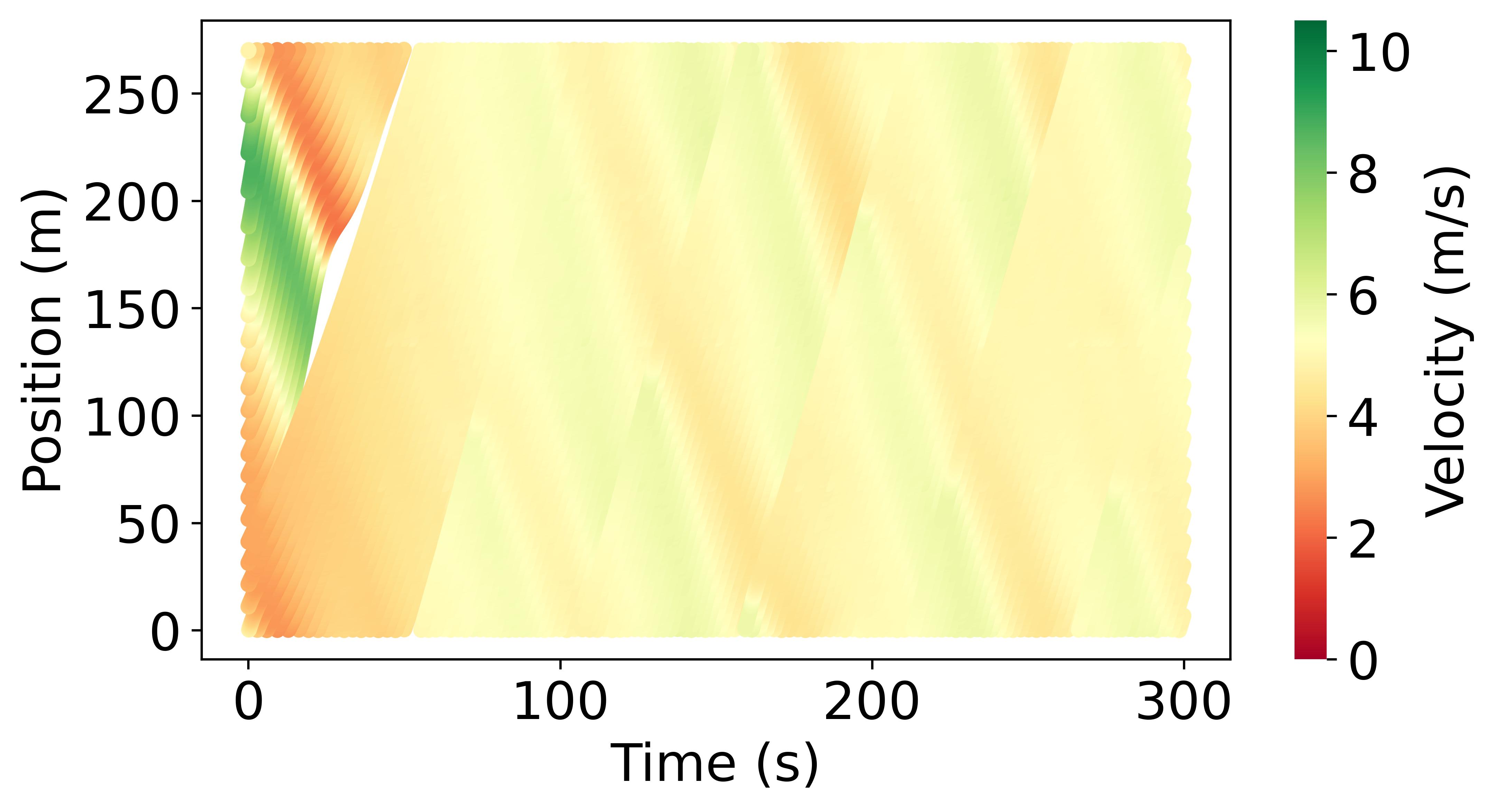}
}
\quad
\subfigure[Figure eight scenario]{
\includegraphics[width=0.441\textwidth,height=4.21cm]{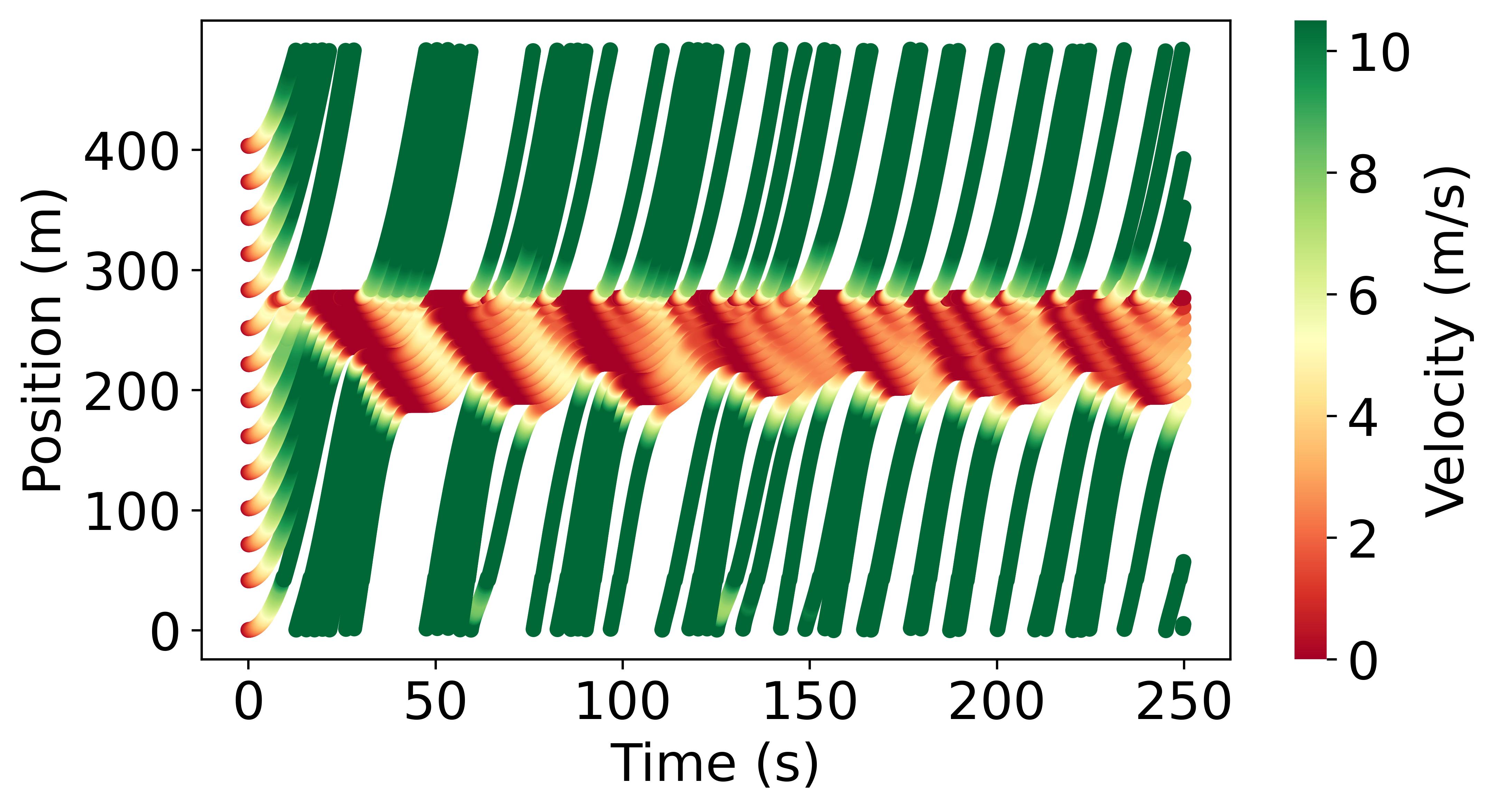}
}
\quad
\subfigure[Figure eight scenario]{
\includegraphics[width=0.441\textwidth,height=4.21cm]{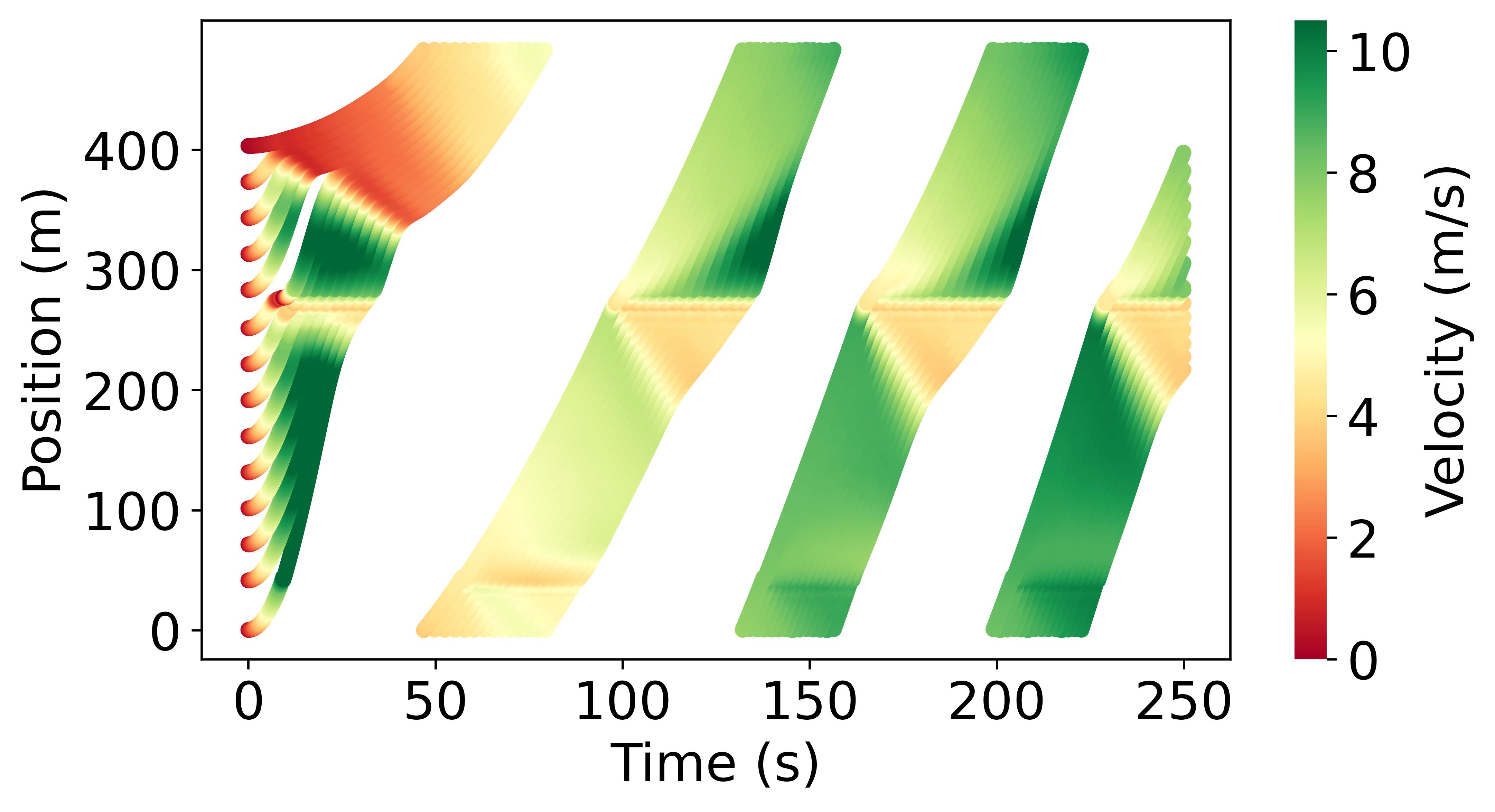}
}
\quad
\subfigure[Merge scenario]{
\includegraphics[width=0.441\textwidth,height=4.21cm]{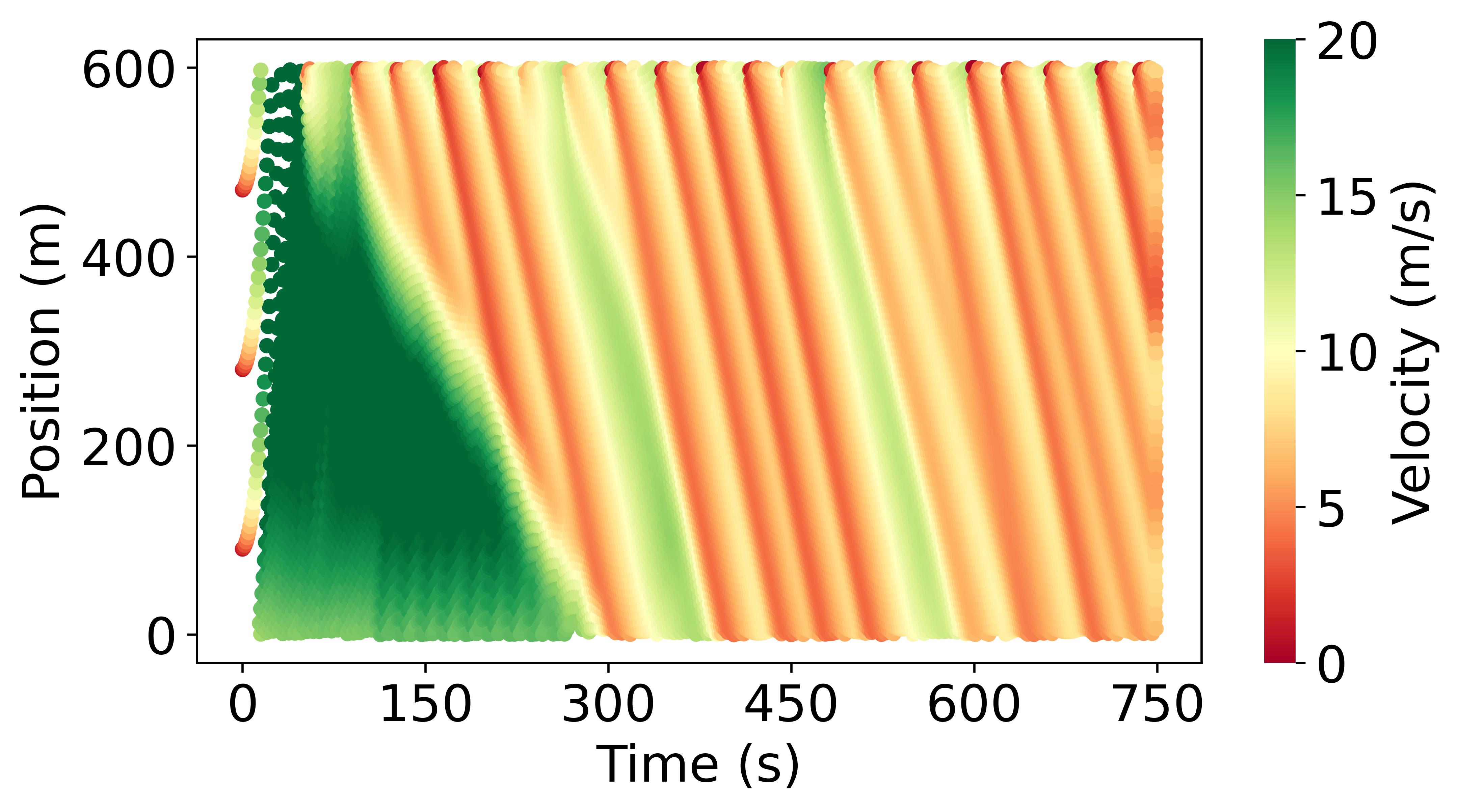}
}
\quad
\subfigure[Merge scenario]{
\includegraphics[width=0.441\textwidth,height=4.21cm]{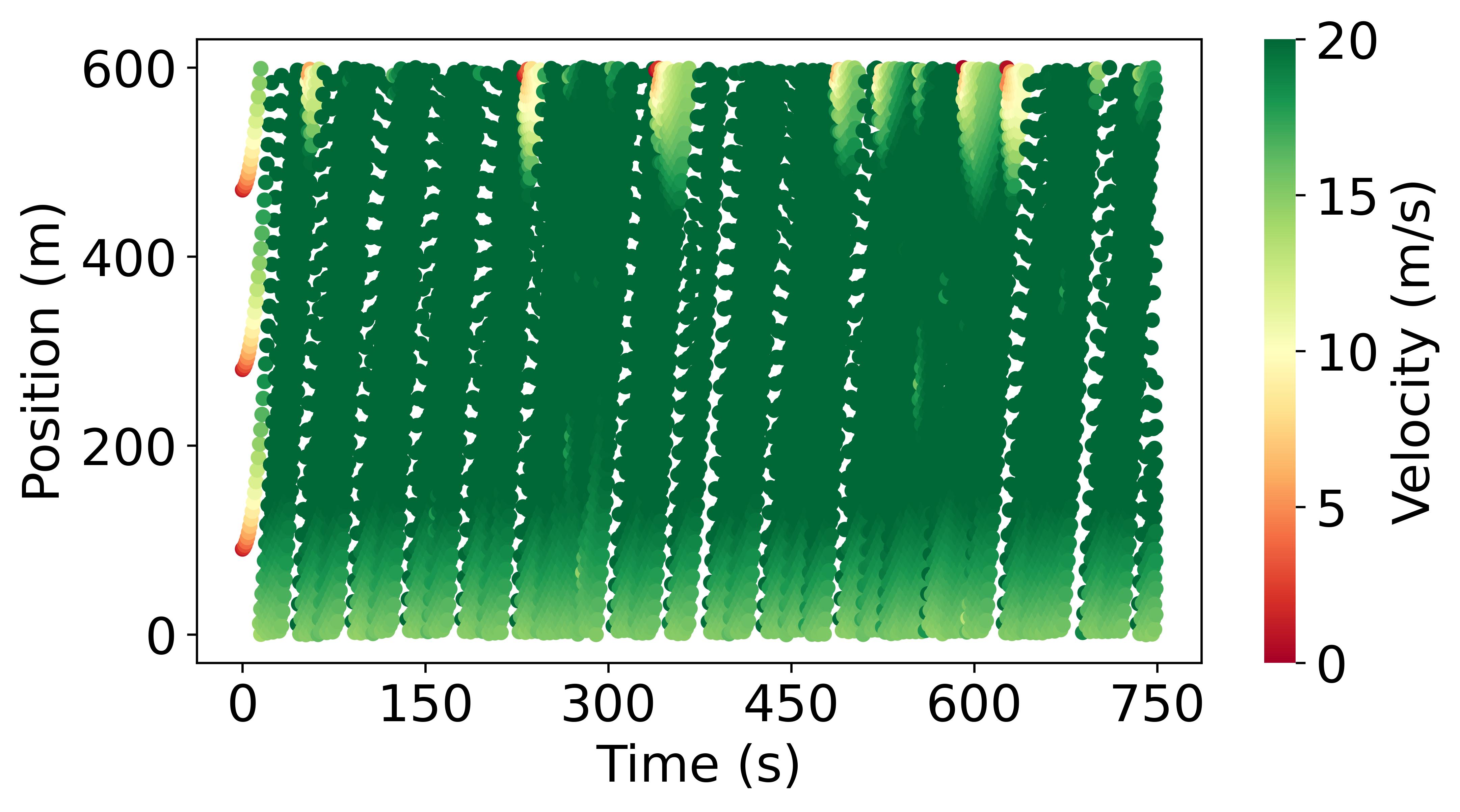}
}
\caption{Space-time trajectories and velocity heat map under different CAV control strategies (IDM and ours) in three scenarios.}
\label{fig5}
\end{figure}

Figure \ref{fig5} presents a comprehensive visual analysis of the space-time trajectories and velocity heat maps under different CAV control strategies in these three traffic scenarios. As shown in Figure \ref{fig5}(a), (c), and (e), when the CAV agent operates under the IDM model, stop-and-go waves persist across all three road networks. These waves manifest as oscillatory patterns in the space-time trajectories, characterized by alternating regions of high and low vehicle speeds. The presence of stop-and-go waves indicates frequent acceleration and deceleration, leading to increased fuel consumption, reduced driving comfort, and overall inefficiencies in the traffic flow. In contrast, when our proposed approach is employed to control the CAV agent, as shown in Figure \ref{fig5}(b), (d), and (f), we can observe smoother and more uniform patterns, indicating a substantial reduction in stop-and-go waves. The velocity heat maps further reinforce this observation, showcasing a more consistent distribution of velocities.

It is worth noting that the gaps observed in Figure \ref{fig5}(c) and (d) are a direct consequence of the unique characteristics of the figure eight network used in our experiments. This network consists of two circular loops connected at opposite ends, forming a closed path resembling an unsignalized intersection. Due to the specific design of this network and the limited number of vehicles it can accommodate (14 in total), there are instances where certain sections of the network are unoccupied by vehicles at specific time steps. These unoccupied sections manifest as gaps in the position and velocity plots. Additionally, the jagged appearance of the trajectories in Figure \ref{fig5}(f) is a result of the higher velocities achieved by the vehicles under the control of the CAVs in the merge network scenario. The jagged patterns in the trajectories arise from the combination of the higher velocities and the discrete nature of the plotted data points. The trajectories are plotted using the positions of vehicles at specific time intervals. As the vehicles move at faster speeds under the control of CAVs, the distance traveled between consecutive data points increases, resulting in more pronounced steps in the trajectory lines. The presence of CAVs in the merge network leads to significantly faster traffic flow compared to the scenario in Figure \ref{fig5}(e), where all vehicles are IDM-controlled. The advanced sensing, communication, and control capabilities of CAVs enable them to optimize their driving behavior, maintain higher speeds, and facilitate smoother merging operations. As a result, the overall traffic flow in the merge network with CAVs is more efficient and experiences less congestion. Overall, Figure \ref{fig5} highlights the superiority of our proposed knowledge-informed model-based residual reinforcement learning approach over the traditional IDM model in mitigating stop-and-go waves and promoting a more efficient and stable traffic state.

\subsubsection{Performance of virtual environment model}
To demonstrate the effectiveness of our proposed traffic knowledge-informed virtual environment modeling, we compare the training loss and prediction accuracy of two models: our proposed approach, denoted as Knowledge NN, and a baseline model, referred to as Vanilla NN. In order to ensure a comprehensive and diverse dataset for training purposes, we utilize the dataset developed by \citet{Mo2021physics}, which encompasses four representative driving scenarios: acceleration, deceleration, cruising, and emergency braking. The deceleration scenario involves accelerations ranging from 0 to -2$\,\mathrm{m/s^2}$, while the emergency braking scenario assumes an initial acceleration of -2$\,\mathrm{m/s^2}$. To further enrich the dataset, the initial gap and velocity of the following vehicle are varied in each scenario. Additionally, the initial speed of the leading vehicle in each simulation is adjusted, and the leading vehicle maintains a constant speed throughout the simulation. To account for real-world stochasticity, Gaussian noise is introduced to the acceleration of the following vehicle. In total, the dataset consists of 262,630 data points, with 70\% randomly selected for training, 10\% for validation, and the remaining 20\% for testing.

\begin{figure*}
\centering
\subfigure[]{
\includegraphics[height=4.55cm]{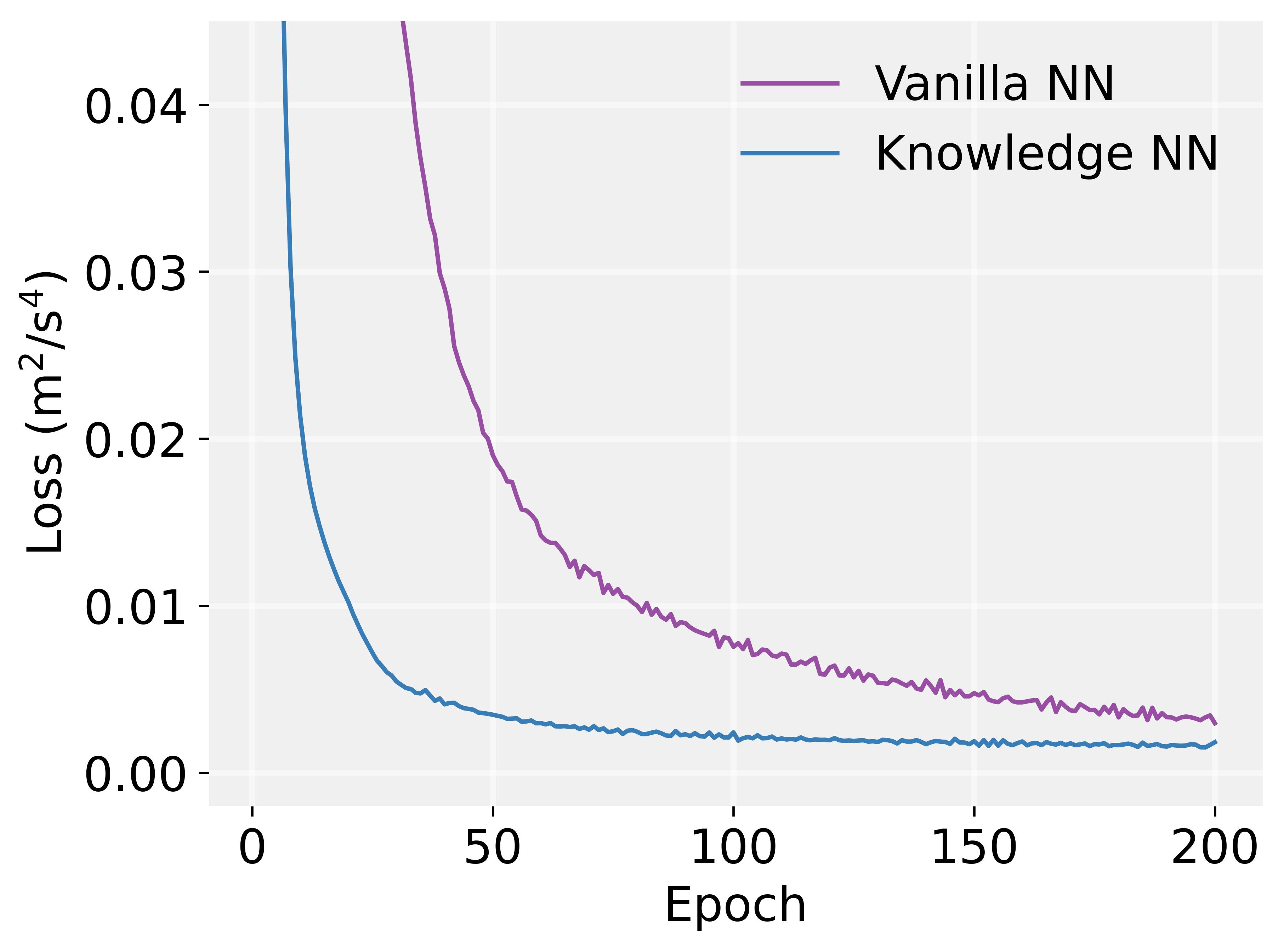}
}
\quad
\subfigure[]{
\includegraphics[height=4.55cm]{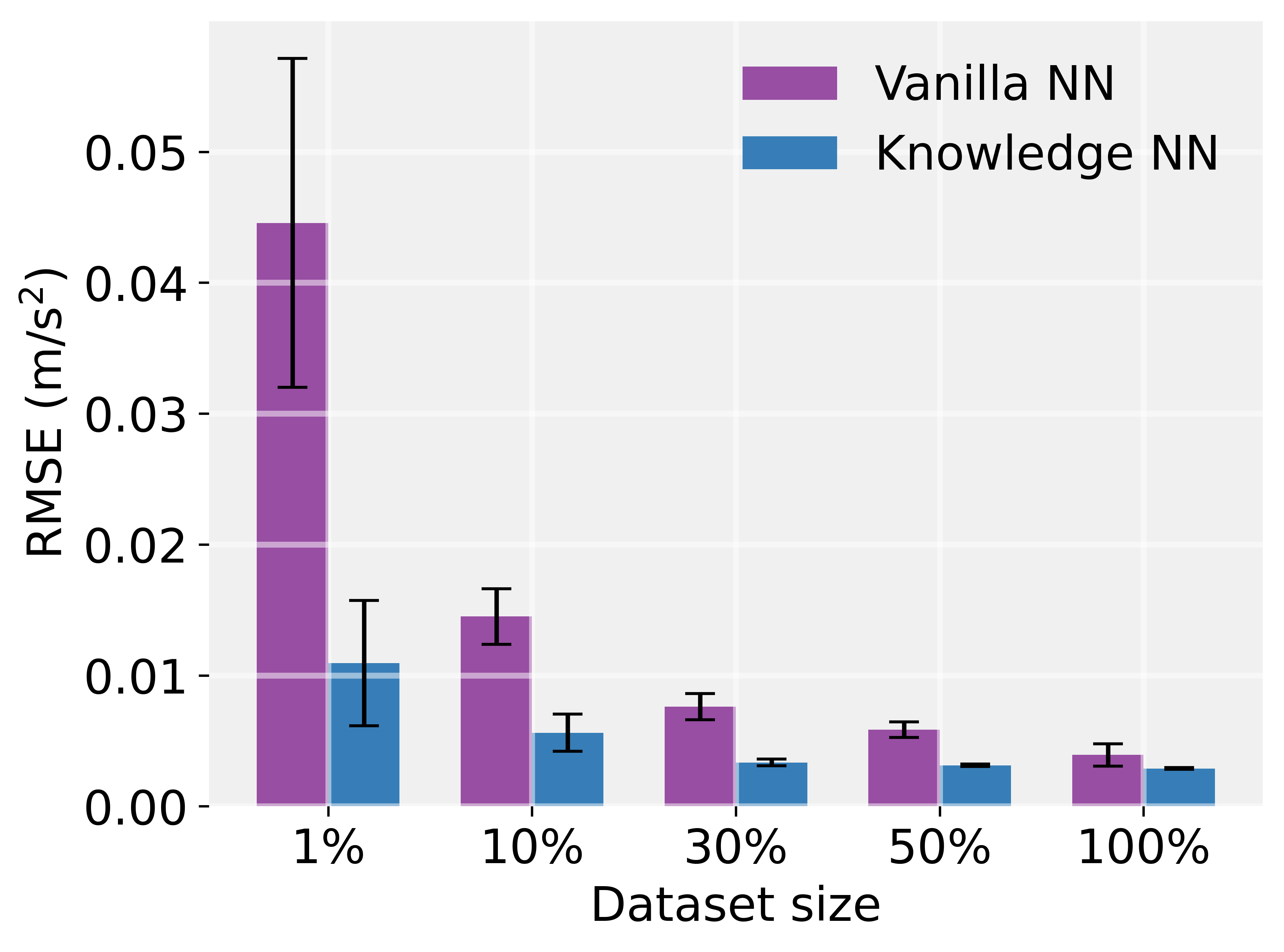}
}
\caption{Comparison of training loss and prediction performance between the Knowledge NN and the Vanilla NN.}
\label{fig6}
\end{figure*}

Figure \ref{fig6}(a) illustrates the training loss curves for both the Knowledge NN and the Vanilla NN. We can find that the Knowledge NN exhibits a significantly faster convergence speed and achieves substantially lower loss values compared to the Vanilla NN. Moreover, Figure \ref{fig6}(b) presents the average prediction error under different training dataset sizes. For each dataset size, we randomly sample a percentage of the data and repeat the process five times, reporting the mean and standard deviation of the prediction error on the testing set. The results indicate that the Knowledge NN consistently outperforms the Vanilla NN in terms of prediction accuracy across all dataset sizes. Furthermore, the Knowledge NN demonstrates more stable prediction performance, as evidenced by the smaller standard deviations. The superior performance of the Knowledge NN can be attributed to the integration of prior knowledge from the IDM. By leveraging domain expertise, the Knowledge NN gains valuable insights to guide the learning process. The combination of expert knowledge and data-driven learning enables Knowledge NN to better capture the underlying system dynamics, resulting in more accurate predictions and more efficient model updates. In contrast, Vanilla NN relies solely on data-driven learning without any prior knowledge, and thus struggles to achieve comparable performance, especially when faced with limited data size. The absence of domain-specific guidance leads to slower convergence, higher loss values, and less accurate predictions. This underscores the importance of incorporating expert knowledge into the virtual environment modeling process, as it can significantly enhance the model ability to learn and generalize from limited data.

\subsubsection{Ablation analysis}
\begin{figure*}
\centering
\subfigure[]{
\includegraphics[height=3.68cm]{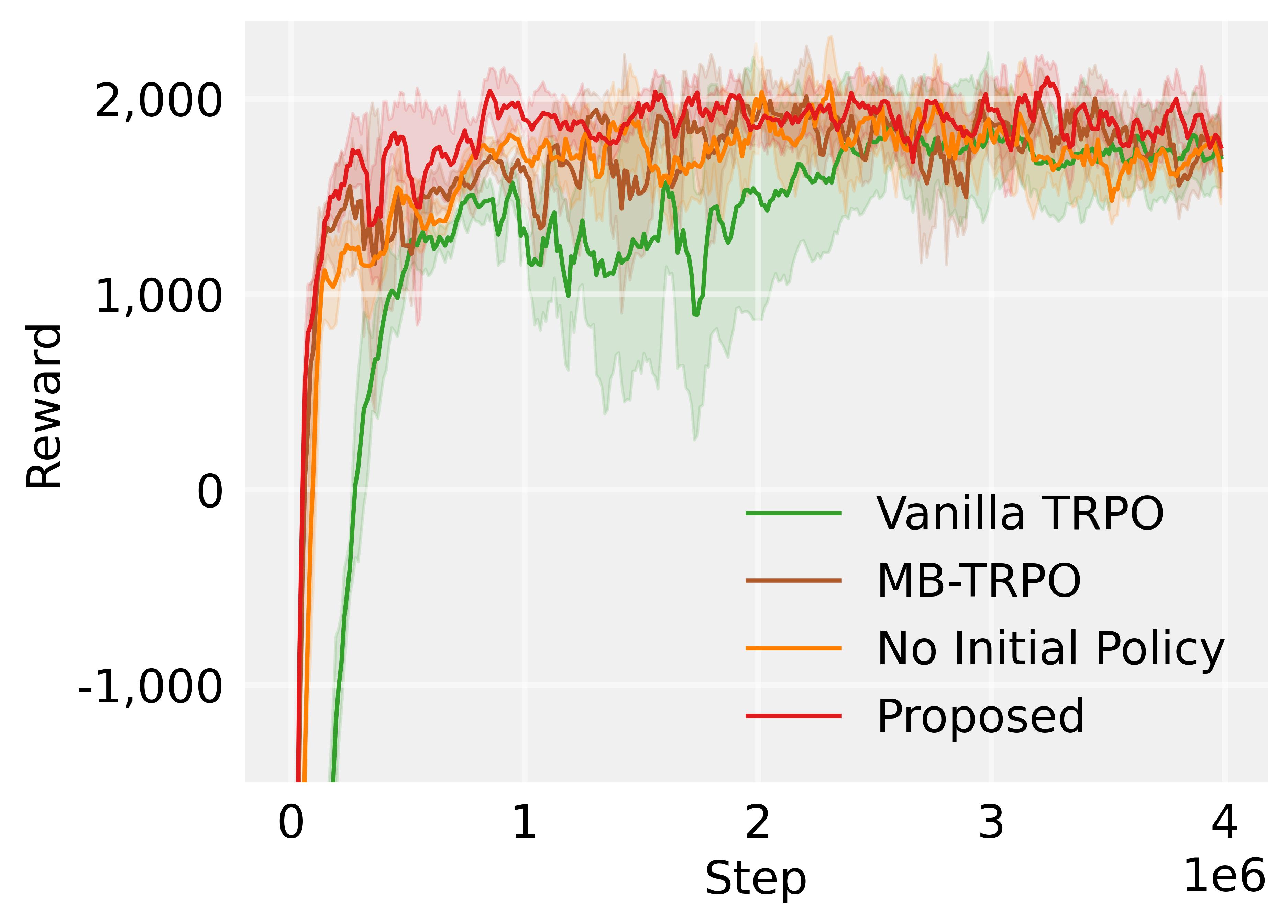}
}
\quad
\subfigure[]{
\includegraphics[height=3.68cm]{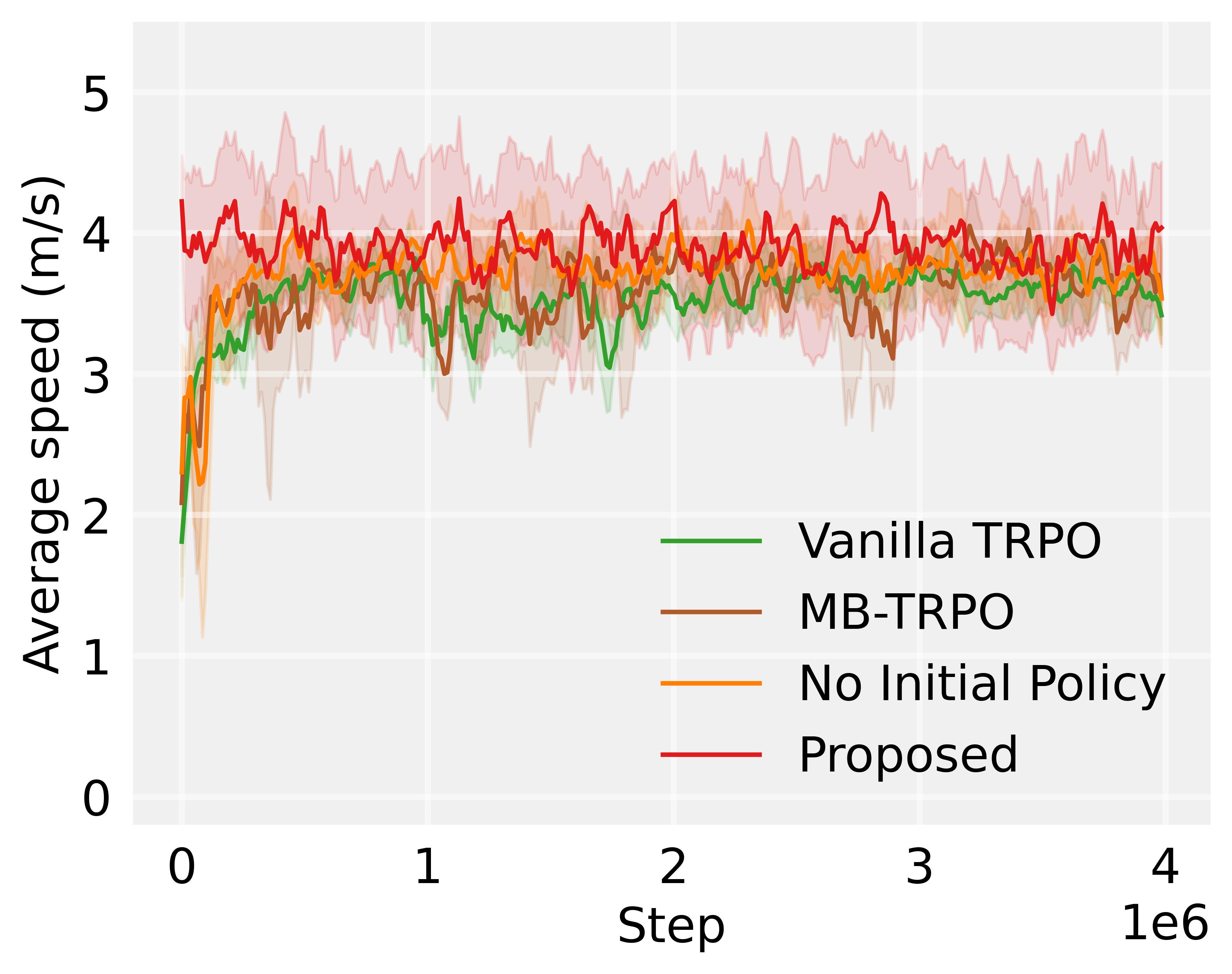}
}
\quad
\subfigure[]{
\includegraphics[height=3.68cm]{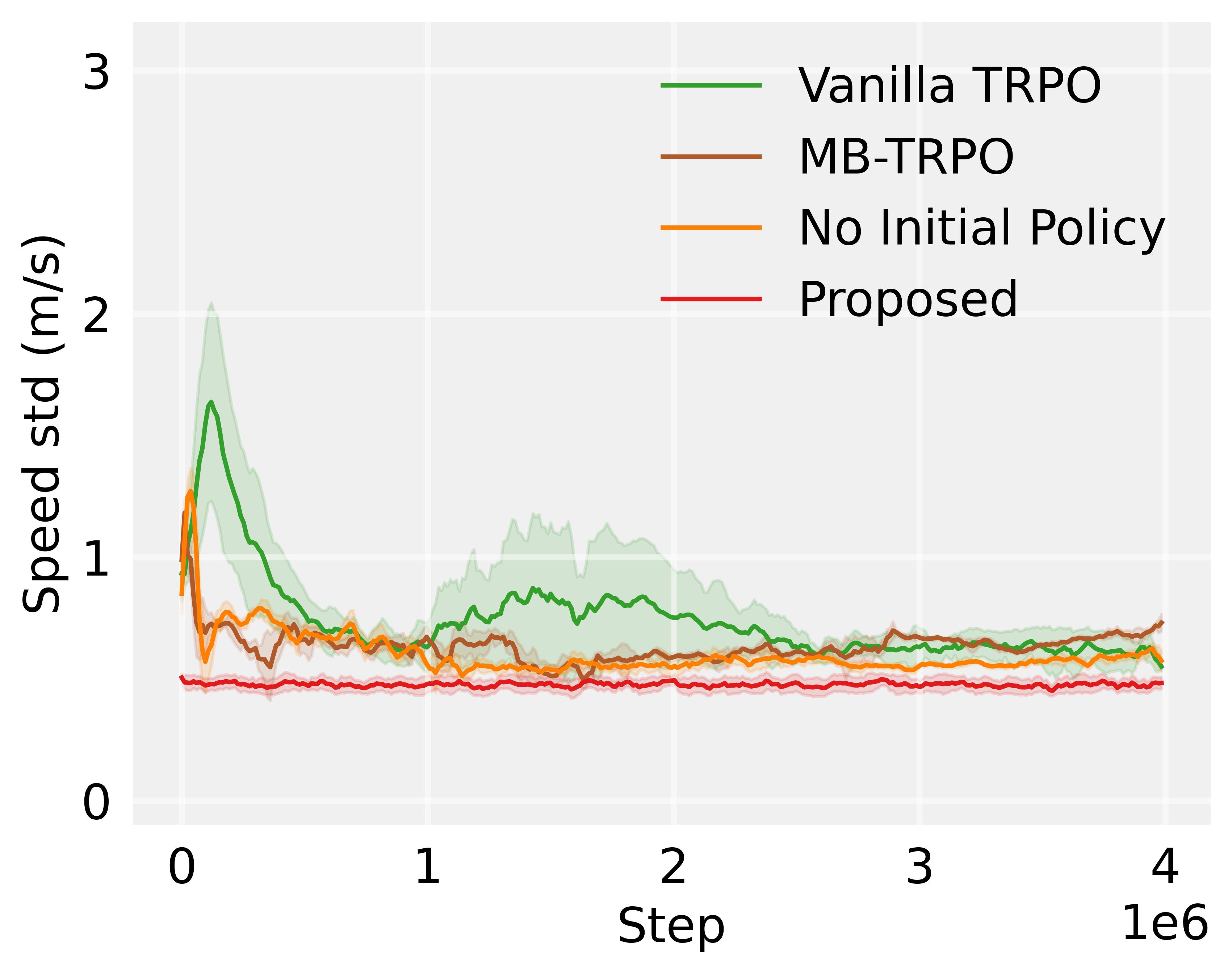}
}
\caption{Ablation analysis of the proposed approach and its variants in the ring scenario.}
\label{fig7}
\end{figure*}

To gain a deeper understanding of the individual contributions of the proposed modules and their synergistic effects, we performed a comprehensive ablation analysis in the ring scenario. The results are presented in Table \ref{tab2} and Figure \ref{fig7}. Specifically, we investigated three variants of our method to isolate the impact of each component. The first variant, denoted as No Initial Policy, excludes the physics-based initial policy from the learning process. This allows us to explore the effect of prior knowledge on learning efficiency and performance. The second variant, referred to as MB-TRPO, further omits the integration of traffic knowledge in the virtual environment modeling, instead relying exclusively on neural networks for model-based RL. This variant enables us to assess the significance of incorporating domain expertise in the construction of an accurate and stable virtual environment model. The third variant, denoted as Vanilla TRPO, serves as a benchmark to evaluate the performance of a purely model-free RL approach without the incorporation of any prior knowledge or virtual environment modeling.

\begin{table}
\centering
\caption{Ablation analysis of our proposed approach at the last training checkpoint.}
\label{tab2}
\small
\begin{tabularx}{\textwidth}{l>{\centering\arraybackslash}X>{\centering\arraybackslash}X>{\centering\arraybackslash}X}
\toprule
{Approach} & {Reward} & {Avg speed} & {Speed std} \\
\midrule
Vanilla TRPO      & 1,699.15 & 3.42 & 0.55 \\
MB-TRPO           & 1,717.45 & 3.55 & 0.73 \\
No Initial Policy & 1,633.28 & 3.53 & 0.57 \\
Proposed          & 1,753.66 & 4.04 & 0.48 \\
\bottomrule
\end{tabularx}
\end{table}

The ablation analysis yields several notable observations. Firstly, our proposed method, which seamlessly integrates the physics-based initial policy, traffic knowledge-informed virtual environment modeling, and residual RL paradigm, achieves the highest reward and fastest convergence among all variants. This underscores the effectiveness of our holistic approach in addressing the complexities of CAV control in mixed traffic environments. Secondly, while the Vanilla TRPO variant achieves a final reward with similar value to others, it exhibits the slowest reward improvement, highlighting the limitations of a purely model-free RL scheme. This observation supports the idea that model-based RL can improve the sample efficiency of RL by utilizing a virtual environment model to generate additional training data. Thirdly, while the MB-TRPO variant demonstrates improved sample efficiency compared to Vanilla TRPO, its reward curve is characterized by fluctuations and instability. This instability can be attributed to the sole reliance on neural networks for learning the environment dynamics. Neural networks may struggle to capture the intricate details and uncertainties present in real-world traffic scenarios. Consequently, the learned virtual environment model may not be sufficiently accurate or stable, leading to suboptimal policy improvements when applied to the actual environment. In contrast, our proposed method leverages the complementary strengths of traffic expert knowledge and data-driven approaches. By incorporating the physics-based initial policy, our method allows the agent to avoid starting the exploration process from scratch, thereby significantly improving learning efficiency. The initial policy, derived from domain expertise, provides a stable foundation for the agent to build upon and refine through the residual reinforcement learning paradigm. Moreover, the integration of traffic knowledge in the virtual environment modeling process leads to a more accurate and stable virtual environment model. The combination of the IDM model, which captures the basic traffic dynamics, and neural networks, which learn the residual dynamics, results in a virtual environment that closely mimics real-world traffic conditions. This enables the agent to learn and optimize policies efficiently and stably. The ablation analysis underscores the importance of each component in our proposed knowledge-informed model-based residual reinforcement learning approach. The synergistic combination of these components leads to superior performance, faster convergence, and more stable learning compared to the variants that lack one or more of these key elements.

\subsubsection{Computational efficiency analysis}
While our primary focus has been on improving traffic flow efficiency and smoothing, we recognize the importance of computational efficiency, particularly in model-based RL approaches. To address this, we conducted a thorough analysis of the computational requirements of our proposed method and compared it with baseline approaches and variant models. All experiments were conducted on a high-performance desktop computer running Ubuntu 20.04, equipped with an Intel Core i9-10980XE CPU, two NVIDIA GeForce RTX 4090 GPUs, and 128GB RAM. This setup ensured a consistent and powerful computational environment for all comparisons.

\begin{figure*}
\centering
\subfigure[]{
\includegraphics[height=4.53cm]{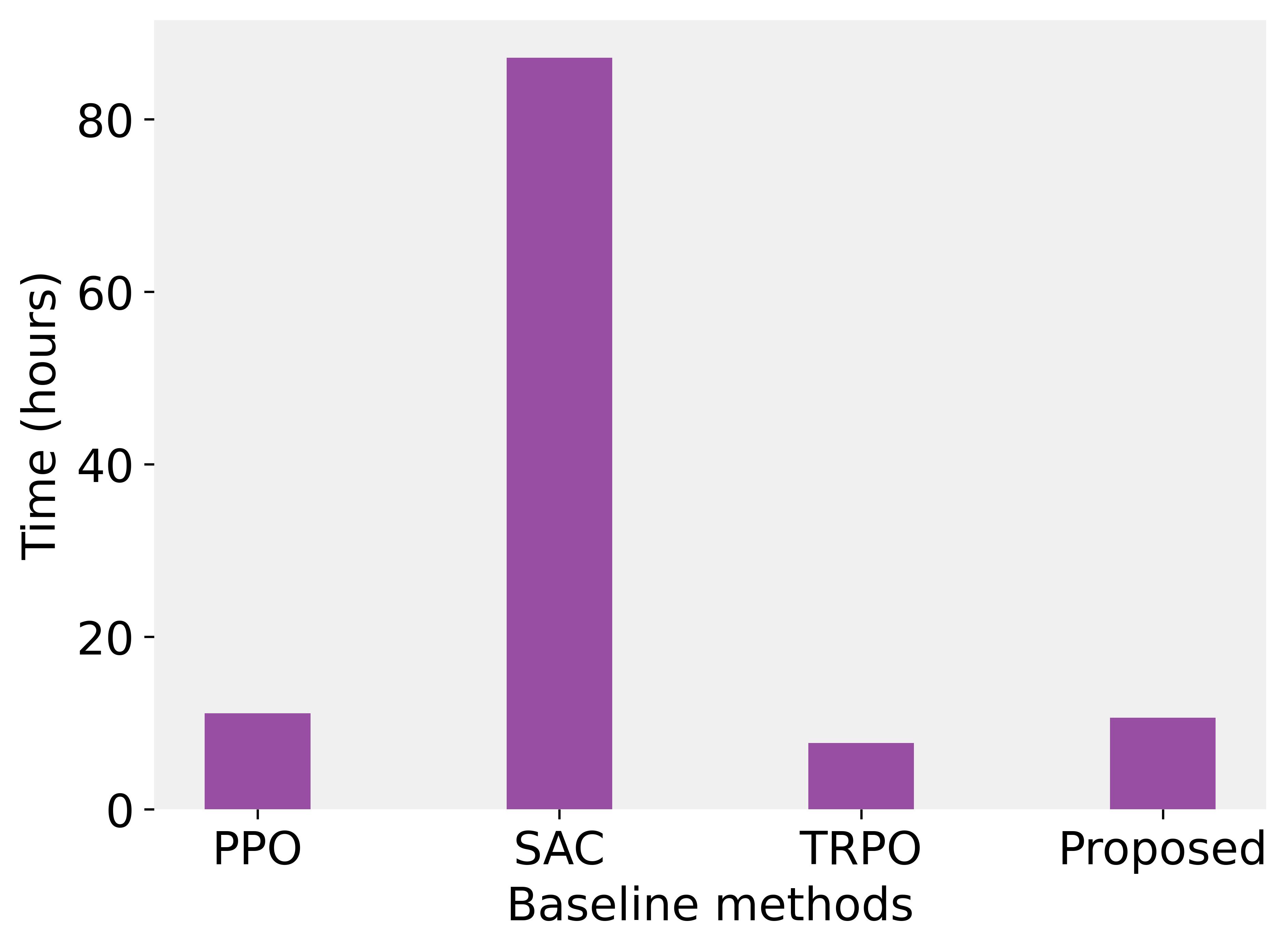}
}
\quad
\subfigure[]{
\includegraphics[height=4.53cm]{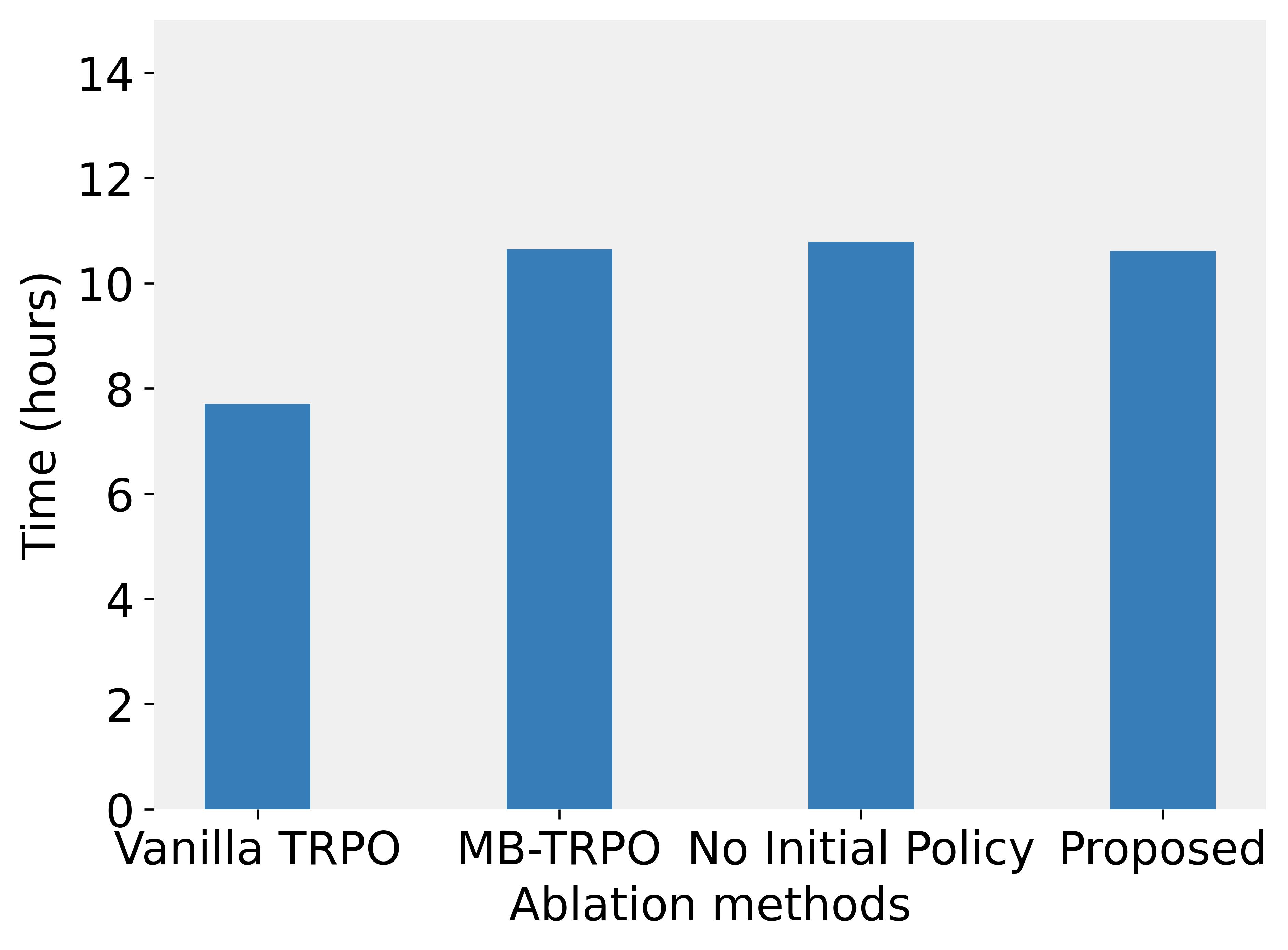}
}
\caption{Comparison of training time among the proposed approach with baseline algorithms and its variants in the ring scenario.}
\label{fig8}
\end{figure*}

Figure \ref{fig8}(a) illustrates the training time required for our proposed method compared to several baseline methods. The proposed method demonstrates favorable computational efficiency compared to PPO and SAC, requiring approximately 4.6\% less training time. However, as shown in Figure \ref{fig8}(b), it does require about 37\% more training time compared to Vanilla TRPO. This increased computational cost can be attributed to the integration of expert knowledge, which provides a more informed starting point for the learning process and constructs a virtual environment model. While this integration introduces additional computational overhead, it significantly enhances the overall learning process. It is important to note that while our method incurs some additional training time, this increase is primarily confined to the offline training phase. Once training is complete, the virtual environment model is no longer necessary for decision-making, significantly reducing the computational overhead during operation. The additional computational cost during training is justified by the performance improvements and enhanced sample efficiency achieved by our approach. The incorporation of expert knowledge, while more computationally intensive, leads to faster convergence, superior performance in complex traffic scenarios, and a more accurate representation of the environmental dynamics.

To further enhance the computational efficiency of this approach, several optimization strategies could be explored in future work. The implementation of parallel computing techniques could leverage the full capabilities of multi-core processors and multiple GPUs. Additionally, the application of model pruning techniques could reduce the complexity of neural networks without sacrificing performance. These optimizations have the potential to significantly reduce training time and computational overhead, making the approach even more efficient and scalable for real-world applications in connected automated vehicles and intelligent transportation systems.

\begin{figure*}
\centering
\subfigure[]{
\includegraphics[height=4.54cm]{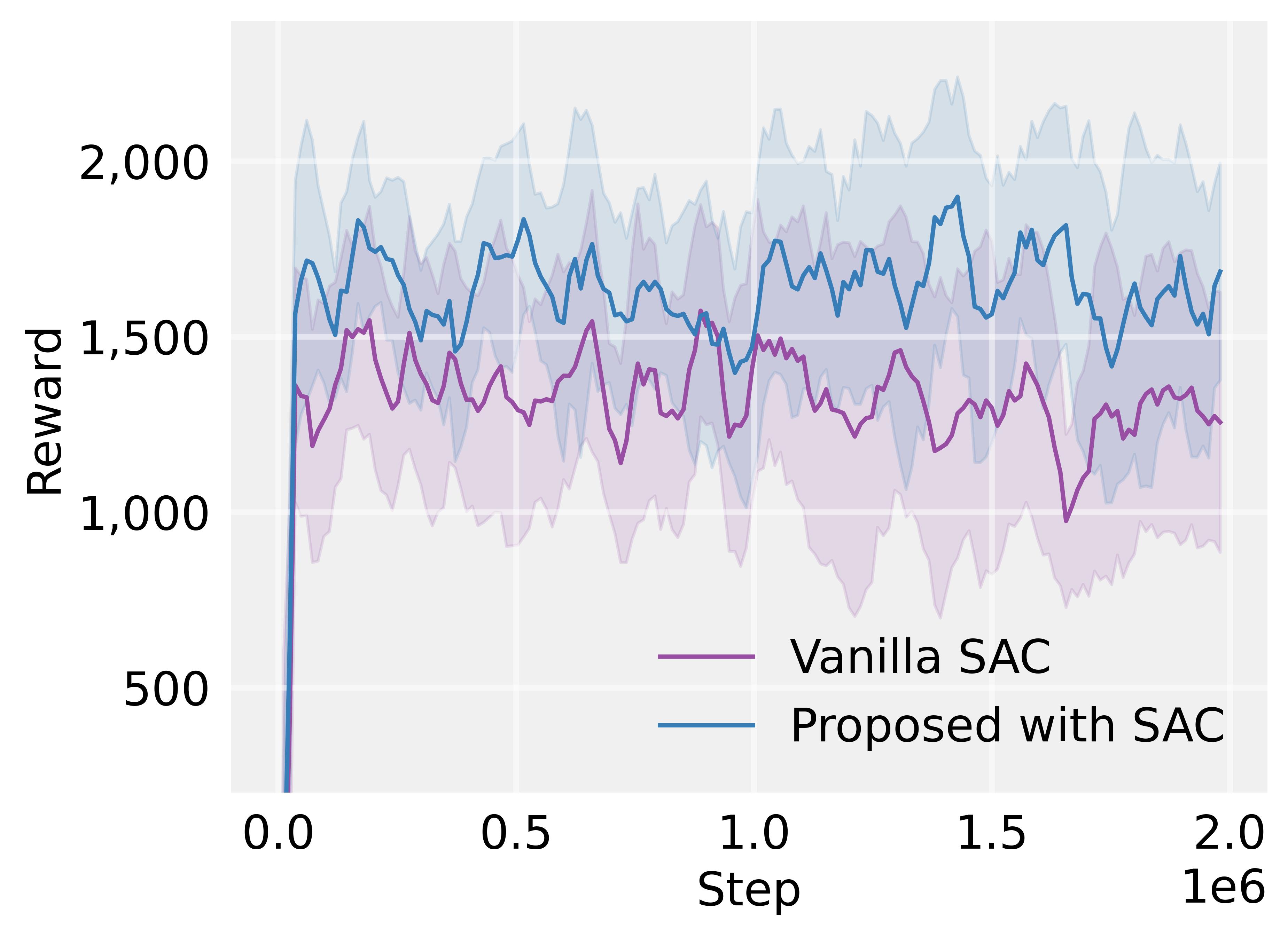}
}
\quad
\subfigure[]{
\includegraphics[height=4.54cm]{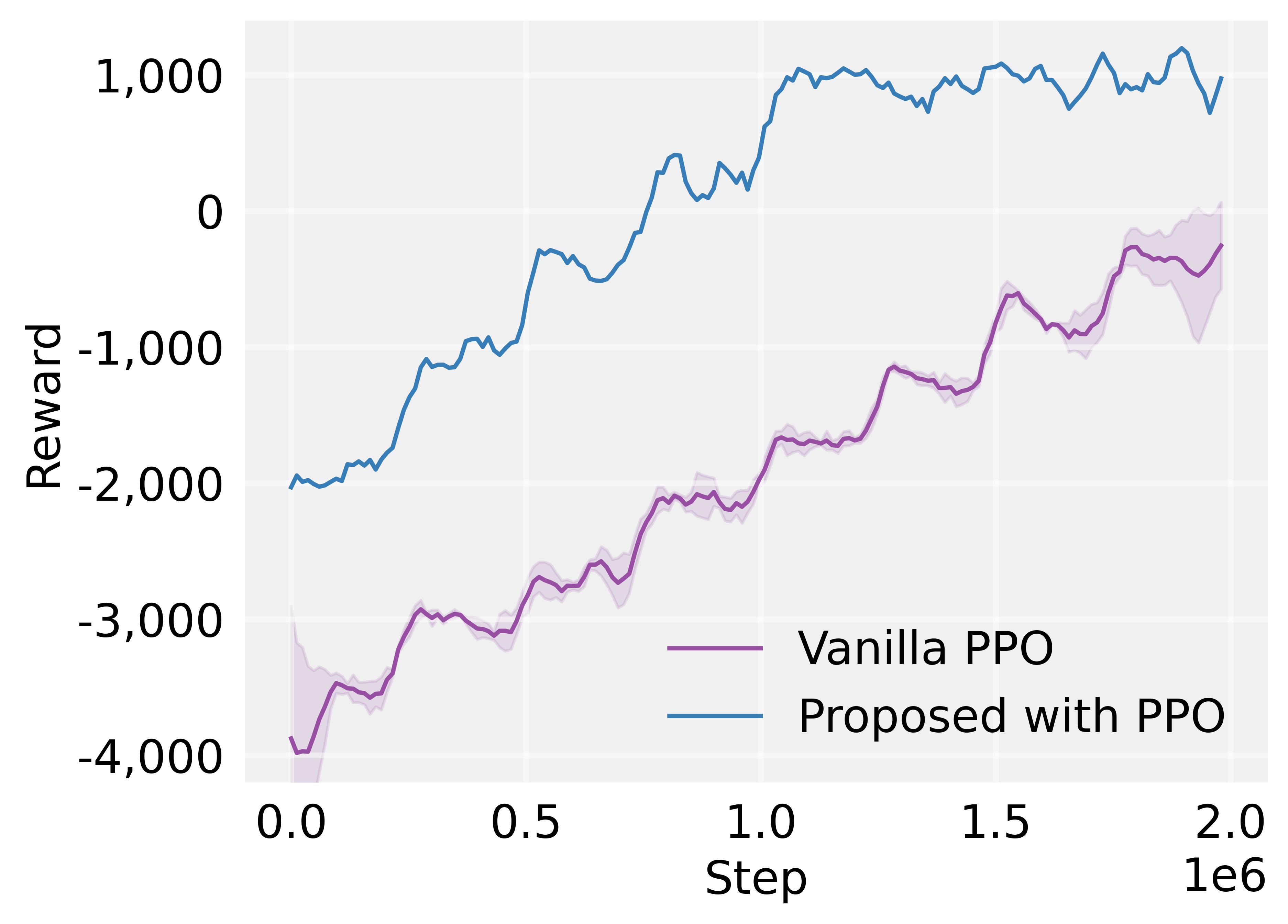}
}
\caption{Sensitivity analysis of the proposed approach using PPO and SAC in the ring scenario.}
\label{fig9}
\end{figure*}   

\subsubsection{Sensitivity analysis}
To demonstrate the flexibility and generalizability of our proposed framework, we conducted additional experiments by replacing the original TRPO algorithm with two other state-of-the-art reinforcement learning algorithms: PPO and SAC. These experiments aimed to assess the framework’s sensitivity to different RL algorithms and its ability to enhance their performance.

Figure \ref{fig9}(a) illustrates the learning curves of our proposed approach integrated with SAC compared to vanilla SAC in the ring road scenario. The results show a consistent performance improvement, with our framework achieving a 34\% increase in reward compared to the vanilla SAC implementation. This enhancement is evident from the higher and more stable reward curve of our proposed approach throughout the training process. Figure \ref{fig9}(b) presents a similar comparison between our framework integrated with PPO and vanilla PPO. The results here are even more striking, with our approach achieving an impressive 483\% increase in reward over the vanilla PPO implementation. The learning curve of our proposed approach with PPO shows a rapid initial improvement and maintains a significantly higher reward level throughout the training process.

These results closely align with the improvements we observed with TRPO in our original experiments, demonstrating the robustness of our framework across different RL algorithms. The consistent performance enhancements across TRPO, PPO, and SAC underscore the versatility of our approach in leveraging expert knowledge and model-based learning to boost the effectiveness of various RL methods in complex traffic scenarios. It is worth noting that the key contribution of our work is not tied to any specific RL algorithm, physical model, or neural network architecture. Rather, we propose a general framework for integrating expert knowledge into model-based reinforcement learning. This framework is designed to be flexible and adaptable to different components. As demonstrated by our experiments with TRPO, PPO, and SAC, our framework can accommodate various RL algorithms, providing a structure for enhancing these algorithms with domain knowledge and residual learning, regardless of their specific update rules or exploration strategies.

\subsubsection{Transferability analysis}

To evaluate the transferability of our proposed approach across different traffic conditions, we conduct a series of experiments with varying traffic volumes in the ring scenario. We consider three distinct volume levels: low volume (5 vehicles, approximately 20 vehicles/km), medium volume (11 vehicles, approximately 40 vehicles/km), and high volume (22 vehicles, approximately 80 vehicles/km). These scenarios are carefully designed to represent a wide range of traffic conditions while maintaining all other parameters constant to isolate the effect of traffic volume.

\begin{figure*}
\centering
\subfigure[Low traffic volume]{
\includegraphics[height=3.51cm]{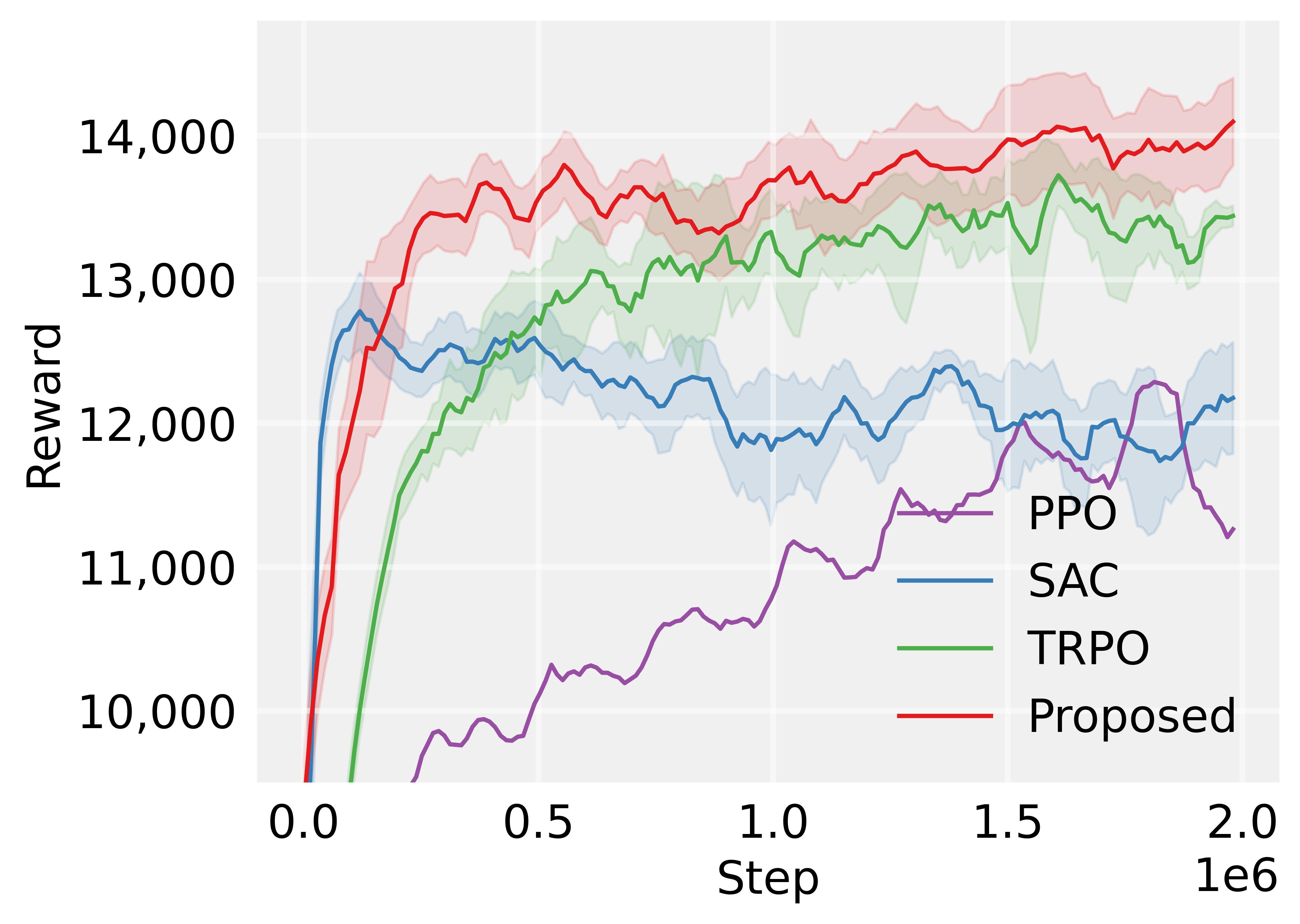}
}
\quad
\subfigure[Medium traffic volume]{
\includegraphics[height=3.51cm]{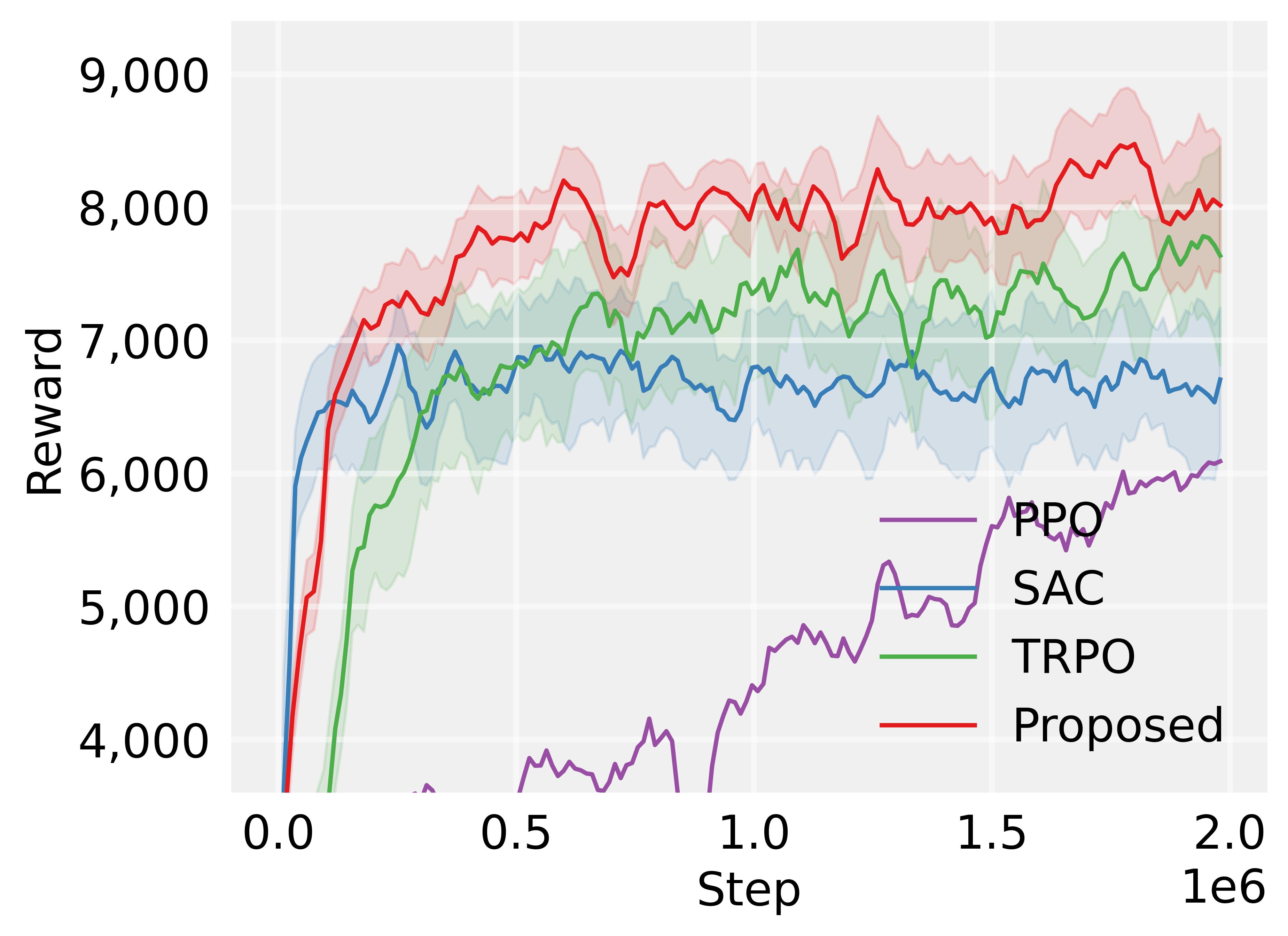}
}
\quad
\subfigure[High traffic volume]{
\includegraphics[height=3.51cm]{fig-ring_returns}
}
\caption{Transferability analysis of the proposed approach with various traffic volumes in the ring scenario.}
\label{fig10}
\end{figure*}

Figure \ref{fig10} shows the results of these experiments, comparing our proposed approach with baseline methods (SAC, PPO, and TRPO) across the different volume levels. The results demonstrate the superior performance and transferability of our method across varying traffic conditions. In the low volume scenario, our approach achieves the highest reward, showcasing its ability to optimize traffic flow even in sparse vehicle conditions. As we transition to medium and high volume scenarios, our method maintains its leading position, consistently outperforming the baseline approaches. This consistent superiority across all volume levels underscores the robustness and transferability of our model. Furthermore, our approach demonstrates superior learning efficiency across all traffic volumes. As evident in Figure \ref{fig10}, our method converges faster and achieves higher reward levels with fewer training steps compared to the baseline methods. Importantly, this high learning efficiency is maintained even as traffic volume increases. The consistent superior performance and high learning efficiency across different traffic volumes have significant implications for the model’s transferability in real-world applications. 

\section{Conclusions}\label{sec5}
In this paper, we proposed a novel knowledge-informed model-based residual reinforcement learning framework for CAVs in mixed traffic environments. Our approach effectively integrates traffic expert knowledge into the reinforcement learning process, enabling the CAV agent to learn and optimize its control policy more efficiently and stably. The proposed framework combines the strengths of model-based RL and domain expertise to address the existing issues of sample inefficiency in model-free RL and the limitation of needing to learn from scratch in model-based RL. By incorporating a physics-based initial policy derived from the PI with saturation controller and a traffic knowledge-informed virtual environment model, our approach significantly improves the sample efficiency and performance of the learning process. The residual reinforcement learning paradigm allows the CAV agent to continuously refine its policy based on the combination of prior knowledge and real-time observations. Extensive experiments conducted in three representative traffic scenarios (ring, figure eight, and merge) demonstrate the superiority of our proposed approach compared to baseline methods. The results show that our approach achieves faster convergence, higher rewards, and more stable learning performance. 

While our current work has shown promising results, there are several exciting directions for future research. First, we aim to explore meta-learning approaches that enable rapid adaptation to new situations and transfer learning techniques that would allow our framework to quickly adapt to new traffic scenarios or regulations without the need for extensive retraining. This could significantly enhance the flexibility and applicability of our system across diverse traffic environments. Second, we plan to bridge the gap between simulation and reality through sim-to-real transfer techniques. Our goal is to conduct small-scale real-world experiments to validate and refine our approach, ensuring its practicality and effectiveness in actual traffic conditions. Third, we intend to expand our framework to multi-agent reinforcement learning scenarios, incorporating additional safety measures for emergent cases. This expansion would allow for more complex and realistic traffic management strategies while prioritizing collision avoidance and overall system safety.

\appendix

\section{Theoretical analysis}
\label{appda}
To provide a theoretical foundation for our proposed approach, we present three key theorems and detailed proofs. These theorems demonstrate the advantages of our physics-based initial policy, the convergence properties of our proposed policy, and the effectiveness of our virtual environment model.

\textbf{Theorem A1.}
Let $\Pi$ be the policy space, $\mathcal{S}$ be the state space, and $\rho_0$ be the initial state distribution. For a physics-based initial policy $\pi_{\text{physics}} \in \Pi$ derived from established transportation science knowledge and a random policy $\pi_{\text{random}} \in \Pi$ initialized from a neural network with random weights, under suitable conditions, there exists $\varepsilon > 0$ such that:
\begin{equation}
\mathbb{E}_{s \sim \rho_0} \left[V^{\pi^*}(s) - V^{\pi_{\text{physics}}}(s)\right] \leq \mathbb{E}_{s \sim \rho_0} \left[V^{\pi^*}(s) - V^{\pi_{\text{random}}}(s)\right] - \varepsilon
\end{equation}
where $\pi^*$ is the optimal policy, and $V^\pi(s)$ is the value function for policy $\pi$.

\textbf{Proof.} We begin by defining $\Delta^\pi(s) = V^{\pi^*}(s) - V^\pi(s)$ as the value difference between the optimal policy and policy $\pi$ at state $s$. Let the expected value difference for a policy $\pi$ be $\mathbb{E}_{s \sim \rho_0} [\Delta^\pi(s)] = \mathbb{E}_{s \sim \rho_0} [V^{\pi^*}(s) - V^\pi(s)]$.

\textbf{Assumption A1.} The physics-based policy $\pi_{\text{physics}}$ is constructed using well-established traffic flow theories and models from transportation science, as well as classical analytical controllers. These models have been validated through extensive empirical studies and are designed to optimize traffic flow and reduce congestion \citep{Stern2018Dissipation, Treiber2000Congested}. In contrast, $\pi_{\text{random}}$ is initialized with random weights, lacking any prior domain knowledge. This random initialization necessitates learning from scratch, resulting in poor initial performance \citep{Huang2024Human}. Therefore, we posit that:
\begin{equation}
\mathbb{E}_{s \sim \rho_0} [V^{\pi_{\text{physics}}}(s)] > \mathbb{E}_{s \sim \rho_0} [V^{\pi_{\text{random}}}(s)]
\end{equation}

From this assumption and the definition of $\Delta^\pi(s)$, we can derive:
\begin{equation}
\mathbb{E}_{s \sim \rho_0} [\Delta^{\pi_{\text{physics}}}(s)] < \mathbb{E}_{s \sim \rho_0} [\Delta^{\pi_{\text{random}}}(s)]
\end{equation}

Therefore, there exists $\varepsilon > 0$ such that:
\begin{equation}
\mathbb{E}_{s \sim \rho_0} [\Delta^{\pi_{\text{physics}}}(s)] \leq \mathbb{E}_{s \sim \rho_0} [\Delta^{\pi_{\text{random}}}(s)] - \varepsilon
\end{equation}

Therefore,
\begin{equation}
\begin{aligned}
\mathbb{E}_{s \sim \rho_0} [V^{\pi^*}(s) - V^{\pi_{\text{physics}}}(s)] &= \mathbb{E}_{s \sim \rho_0} [\Delta^{\pi_{\text{physics}}}(s)] \\
&\leq \mathbb{E}_{s \sim \rho_0} [\Delta^{\pi_{\text{random}}}(s)] - \varepsilon \\
&= \mathbb{E}_{s \sim \rho_0} [V^{\pi^*}(s) - V^{\pi_{\text{random}}}(s)] - \varepsilon
\end{aligned}
\end{equation}

This completes the proof, demonstrating that the physics-based initial policy $\pi_{\text{physics}}$ provides a better starting point closer to the optimal policy $\pi^*$ compared to a random policy $\pi_{\text{random}}$, with the improvement quantified by $\varepsilon$.

\textbf{Theorem A2.}
Under suitable learning conditions, the proposed policy $\pi = \pi_{\text{physics}} + \pi_{\text{residual}}$, where $\pi_{\text{physics}}$ is a physics-based initial policy and $\pi_{\text{residual}}$ is a learned residual policy, can converge to an $\epsilon$-optimal policy $\pi^*_\epsilon$ such that $\|\pi^*_\epsilon - \pi^*\|_\infty < \epsilon$ for any $\epsilon > 0$, where $\pi^*$ is the optimal policy.

\textbf{Proof.} The key to our proof lies in the representation power of $\pi_{\text{residual}}$. Let $\pi_{\text{residual}}$ be represented by a neural network with sufficient capacity. By the Universal Approximation Theorem \citep{Hornik1989Multilayer}, for any continuous function $f$ on a compact subset of $\mathbb{R}^n$ and $\epsilon > 0$, there exists a neural network that can approximate $f$ with an error less than $\epsilon$.

Consider the function $g(s,a) = \pi^*(a \mid s) - \pi_{\text{physics}}(a \mid s)$. There exists a neural network representation $\pi^*_{\text{residual}}$ such that: 
\begin{equation}
\|\pi^*_{\text{residual}}(a \mid s) - g(s,a)\|_\infty < \frac{\epsilon}{2}
\end{equation}
This implies: 
\begin{equation}
\|\left(\pi_{\text{physics}} + \pi^*_{\text{residual}} \right) - \pi^*\|_\infty < \frac{\epsilon}{2}
\end{equation}
Now, we need to show that our learning process can converge to $\pi^*_{\text{residual}}$. We use the following lemma:

\textbf{Lemma A1.} Under suitable conditions (e.g., appropriate learning rate schedule, sufficient exploration), policy gradient methods converge to a local optimum of the expected discounted return $J(\theta)$ \citep{Agarwal2020Optimality}.

In our case, due to the representation power of the neural network, this local optimum corresponds to a policy $\pi^*_\epsilon$ that satisfies:
\begin{equation}
\|\pi^*_\epsilon - (\pi_{\text{physics}} + \pi^*_{\text{residual}})\|_\infty < \frac{\epsilon}{2}
\end{equation}
By the triangle inequality:
\begin{equation}
\begin{aligned}
\|\pi^*_\epsilon - \pi^*\|_\infty &\leq \|\pi^*_\epsilon - (\pi_{\text{physics}} + \pi^*_{\text{residual}})\|_\infty \\
&+ \|(\pi_{\text{physics}} + \pi^*_{\text{residual}}) - \pi^*\|_\infty \\
&< \frac{\epsilon}{2} + \frac{\epsilon}{2} \\
&= \epsilon
\end{aligned}
\end{equation}
Therefore, we have proved that our proposed policy $\pi = \pi_{\text{physics}} + \pi_{\text{residual}}$ can converge to an $\epsilon$-optimal policy $\pi^*_\epsilon$ for any $\epsilon > 0$.

\textbf{Theorem A3.}
Let $T$ be the true environment’s transition dynamics and $\hat{T}$ be our virtual environment model, where $\hat{T} = T_{\text{physics}} + T_{\text{residual}}$. $T_{\text{physics}}$ is derived from a physics-based model, and $T_{\text{residual}}$ is a learned residual component. Under certain conditions, policy optimization using this virtual environment model leads to effective policy improvement in the true environment.

\textbf{Proof.} We begin by defining the model error bound. Let $\epsilon_m$ be the maximum total variation distance between the transition dynamics of the virtual environment model and the true environment:
\begin{equation}
\epsilon_m = \max_t \mathbb{E}_{\tau \sim \pi} \left[ D_{\text{TV}} \left[ T(s, a) \parallel \hat{T}(s, a) \right] \right]
\end{equation}

Following Theorem from \citet{Janner2019When}, we can bound the performance of a policy $\pi$ in the true environment:
\begin{equation}
\eta(\pi) \geq \hat{\eta}(\pi) - 2R_{\text{max}} \left[ \frac{\gamma^{k+1} \epsilon_\pi}{(1-\gamma)^2} + \frac{\gamma^k \epsilon_\pi}{1-\gamma} + \frac{k\epsilon_m}{1-\gamma} \right]
\end{equation}
where $\eta(\pi)$ and $\hat{\eta}(\pi)$ denote the expected returns of policy $\pi$ in the true and virtual environments respectively, and $\epsilon_\pi$ denotes the bounded total-variation distance due to the policy update.

The key insight of our approach is that by learning $T_{\text{residual}}$, we can progressively reduce $\epsilon_m$. We hypothesize that as training progresses:
\begin{equation}
\lim_{t \to \infty} D_{\text{TV}} \left[ T_{\text{residual}} \parallel \hat{T} - T_{\text{physics}} \right] = \lim_{t \to \infty} D_{\text{TV}} \left[ T_{\text{residual}} \parallel T - T_{\text{physics}} \right]
\end{equation}
This implies that $\epsilon_m$ will decrease over training time, tightening the lower bound on $\eta(\pi)$. Additionally, as stated in Assumption A1, the advantage of using $T_{\text{physics}}$ based on domain knowledge (e.g., IDM) is that it can provide a better initial estimation compared to purely data-driven approaches. We formalize this as:
\begin{equation}
D_{\text{TV}} \left[ T_{\text{physics}}(s, a) \parallel T(s, a) \right] \leq D_{\text{TV}} \left[ T_{\text{random}}(s, a) \parallel T(s, a) \right]
\end{equation}
where $T_{\text{random}}$ is a randomly initialized neural network model. This theoretical analysis supports that our virtual environment modeling with residual model approach, by combining $T_{\text{physics}}$ and $T_{\text{residual}}$, can start from a potentially better initial point and continuously reduce the model error $\epsilon_m$ as training progresses. Therefore, we have proved that under certain conditions, policy optimization using our virtual environment model indeed leads to effective policy improvement in the true environment.

\section{Hyperparameters}
\label{appdb}

\setcounter{table}{0}

\begin{table}[H]
\centering
\caption{Parameters of IDM.}
\label{tabb1}
\small
\begin{tabular}{ll}
\toprule
{Parameters} & {Values} \\
\midrule
Desired velocity $v_0$     & 30$\,\mathrm{m/s}$ \\
Desired time headway $T_0$ & 1$\,\mathrm{s}$ \\
Maximum acceleration $a_{\max}$ & 1$\,\mathrm{m/s^2}$ \\
Comfortable deceleration $b$   & 1.5$\,\mathrm{m/s^2}$ \\
Acceleration exponent $\delta$ & 4 \\
Minimum desired gap $s_0$ & 2$\,\mathrm{m}$ \\
Noise & $\mathcal{N}$(0,0.2) \\
\bottomrule
\end{tabular}
\end{table}

\begin{table}[H]
\centering
\caption{Parameters of PI with saturation.}
\label{tabb2}
\small
\begin{tabular}{ll}
\toprule
{Parameters} & {Values} \\
\midrule
Maximum additional velocity  $v^\text{c}$     & 1$\,\mathrm{m/s}$ \\
Lower threshold  $s_l$ & 7$\,\mathrm{m}$ \\
Upper  threshold  $s_u$ & 30$\,\mathrm{m}$ \\
Safety distance   $\Delta x^\text{s}$ & 4$\,\mathrm{m}$ \\
\bottomrule
\end{tabular}
\end{table}

\begin{table}[H]
\centering
\caption{Parameters of PPO.}
\label{tabb3}
\small
\begin{tabular}{ll}
\toprule
{Parameters} & {Values} \\
\midrule
Discount factor $\gamma$ & 0.99 \\
Policy epochs in one PPO update & 80 \\
Clip parameter & 0.2 \\
Actor learning rate & 3$\mathrm{e}$-4 \\
Critic learning rate & 1$\mathrm{e}$-3 \\
\bottomrule
\end{tabular}
\end{table}

\begin{table}[H]
\centering
\caption{Parameters of SAC.}
\label{tabb4}
\small
\begin{tabular}{ll}
\toprule
{Parameters} & {Values} \\
\midrule
Discount factor $\gamma$ & 0.99 \\
Target smoothing coefficient & 5$\mathrm{e}$-3 \\
Learning rate & 3$\mathrm{e}$-4 \\
Temperature parameter & 0.2 \\
Start steps & 1$\mathrm{e}$4 \\
Batch size & 256 \\
\bottomrule
\end{tabular}
\end{table}

\begin{table}[H]
\centering
\caption{Parameters of TRPO.}
\label{tabb5}
\small
\begin{tabular}{ll}
\toprule
{Parameters} & {Values} \\
\midrule
Discount factor $\gamma$ & 0.995 \\
GAE parameter & 0.97 \\
L2 regularization regression & 1$\mathrm{e}$-3 \\
Max KL value $\delta$ & 1$\mathrm{e}$-2 \\
\bottomrule
\end{tabular}
\end{table}

\begin{table}[H]
\centering
\caption{Parameters of our proposed approach.}
\label{tabb6}
\small
\begin{tabular}{ll}
\toprule
{Parameters} & {Values} \\
\midrule
Discount factor $\gamma$ & 0.995 \\
Max KL value $\delta$ & 1$\mathrm{e}$-2 \\
Upper bound for rollout step $k_{\max}$ & 5$\mathrm{e}$2 \\
Rollout length sensitivity $\kappa$ & 2 \\
Desired velocity $v_\text{des}$ & 30$\,\mathrm{m/s}$ \\
Headway threshold $h_{\max}$ & 1$\,\mathrm{s}$ \\
Weighting factor $\alpha$ & 1 \\
Weighting factor $\beta$ & 0.1 \\
Weighting factor $\gamma$ & 0.1 \\
Action & $\pm 1\,\mathrm{m/s^2}$ \\
\bottomrule
\end{tabular}
\end{table}

\bibliographystyle{elsarticle-harv} \biboptions{authoryear}
\bibliography{commtr}

\end{document}